\documentclass{article}
\usepackage{ijcai23}

\usepackage{times}
\usepackage{soul}
\usepackage{url}
\usepackage[hidelinks]{hyperref}
\usepackage[utf8]{inputenc}
\usepackage[small]{caption}
\usepackage{graphicx}
\usepackage{amsmath}
\usepackage{amsthm}
\usepackage{booktabs}
\usepackage{algorithm}
\usepackage{algorithmic}
\usepackage[switch]{lineno}

\usepackage{subcaption}
\usepackage{booktabs}

\usepackage{appendix}
\usepackage{color}
\usepackage{soul}

\title{Shaken, and Stirred: \\Long-Range Dependencies Enable Robust Outlier Detection with PixelCNN++}

\author{
Barath Mohan Umapathi$^1$
\and
Kushal Chauhan$^2$\and
Pradeep Shenoy$^2$\And
Devarajan Sridharan$^{3,4,*}$
\affiliations
$^1$Department of Physics, Indian Institute of Science\\
$^2$Google Research\\
$^3$Center for Neuroscience, Indian Institute of Science\\
$^4$Computer Science and Automation, Indian Institute of Science
\emails
barathu@iisc.ac.in,
\{kushalchauhan, shenoypradeep\}@google.com,
sridhar@iisc.ac.in
}

\begin{document}

\maketitle

\begin{abstract}
Reliable outlier detection is critical for real-world deployment of deep learning models. Although extensively studied, likelihoods produced by deep generative models have been largely dismissed as being impractical for outlier detection. First, deep generative model likelihoods are readily biased by low-level input statistics. Second, many recent solutions for correcting these biases are computationally expensive, or do not generalize well to complex, natural datasets. Here, we explore outlier detection with a state-of-the-art deep autoregressive model: PixelCNN++. We show that biases in PixelCNN++ likelihoods arise primarily from predictions based on local dependencies. We propose two families of bijective transformations -- ``stirring'' and ``shaking'' -- which ameliorate low-level biases and isolate the contribution of long-range dependencies to PixelCNN++ likelihoods. These transformations are inexpensive and readily computed at evaluation time. We test our approaches extensively with five grayscale and six natural image datasets and show that they achieve or exceed state-of-the-art outlier detection, particularly on datasets with complex, natural images. We also show that our solutions work well with other types of generative models (generative flows and variational autoencoders) and that their efficacy is governed by each model's reliance on local dependencies. In sum, lightweight remedies suffice to achieve robust outlier detection on image data with deep generative models.\footnote{Code associated with this paper is available at: 
\url{https://github.com/coglabiisc/googleresearch/tree/main/pixelcnn_ood}\\
\hspace*{1.6em}* Corresponding author}
\end{abstract}

\section{Introduction}

Deep discriminative models confidently misclassify test samples far removed from their training distributions \cite{hendrycks2017a}. By contrast, deep generative models (DGMs) offer a potentially promising approach for identifying outliers. DGMs model the likelihood distribution of the in-distribution (ID) training data and should, in principle, assign lower likelihoods to unfamiliar, out-of-distribution (OOD) samples. In practice, however, DGMs can assign higher likelihoods to OOD samples because of biases arising from low-level image statistics~\cite{Ren2019,Nalisnick2019,Choi2018,Xiao2020}.

\begin{figure*}[tbh]
\centering
\includegraphics[width=0.85\linewidth]{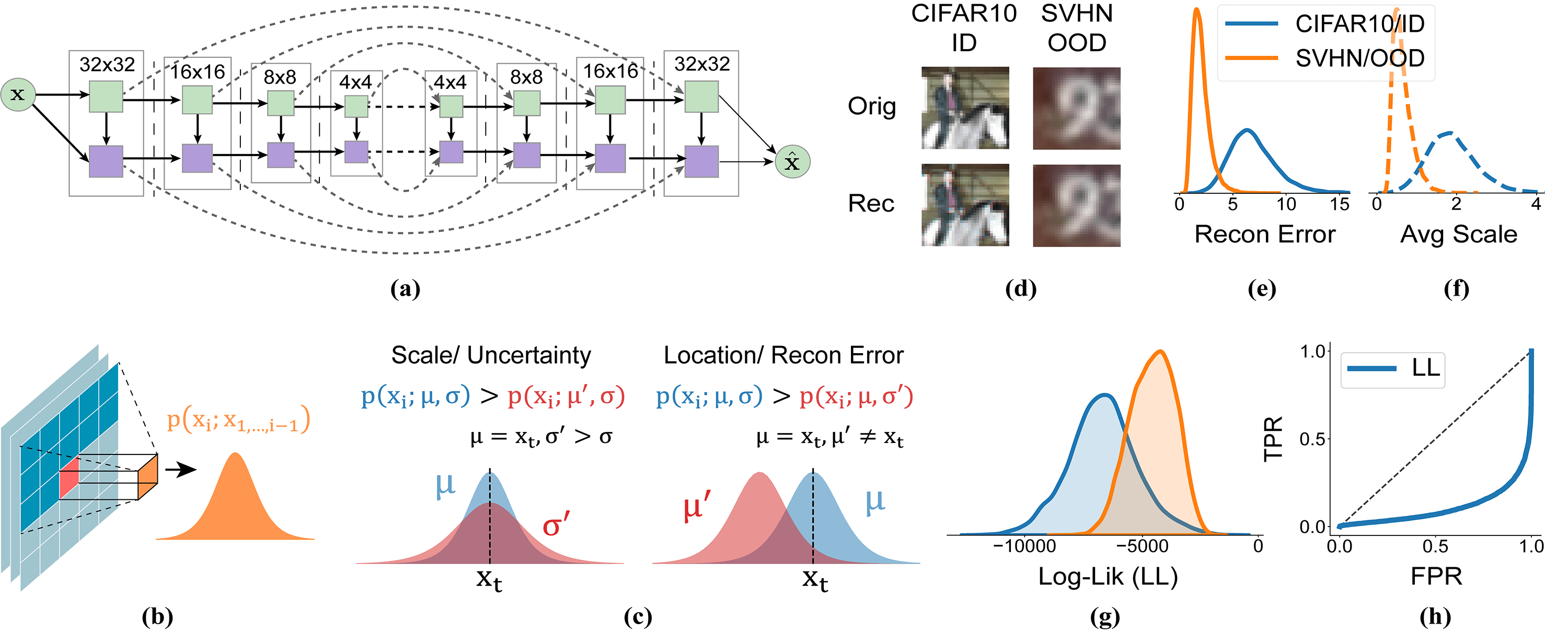}
\caption{\textbf{Biases with PixelCNN++  likelihoods. (a)} PixelCNN++ model architecture. We use one more hierarchy, including a nested set of $4 \times 4$ convolutional layers, compared to the original PixelCNN++ model. Green and purple blocks: vertical and horizontal stacks, respectively. Solid arrows: convolutional connections; curved, dashed arrows: short-cut connections. \textbf{(b)} PixelCNN++ models the likelihood of the current pixel $x_i$ based on the preceding rows and columns of pixels.  \textbf{(c)} (Left) Larger the scale of the predicted logistic, the higher the uncertainty, and the lower the likelihood for the target pixel $\mathbf{x}_t$. (Right) The higher the deviation between the mode of the predicted logistic and $\mathbf{x}_t$, the higher the reconstruction error, and the lower the likelihood for the target pixel. \textbf{(d)} PixelCNN++ model trained on the CIFAR10 datasets reconstructs both CIFAR10/ID and SVHN/OOD samples well. \textbf{(e-f)} PixelCNN++ reconstruction error (e) and average predicted scale (f) distributions for CIFAR10/ID (blue) and SVHN/OOD (orange) test samples. Surprisingly, SVHN/OOD samples have lower reconstruction error and scale as compared to CIFAR10/ID. \textbf{(g)} PixelCNN++ log likelihood distributions. SVHN/OOD samples get higher log-likelihoods than CIFAR10/ID samples. \textbf{(h)} ROC curve for outlier detection using PixelCNN++ likelihoods for CIFAR10/ID and SVHN/OOD. The curve is bowed downwards, indicating anomalously higher likelihoods for OOD than ID samples.}
\label{fig1}
\vspace{-1em}
\end{figure*}

Here, we analyze biases in likelihoods produced by a state-of-the-art DGM: PixelCNNs~\cite{Oord2016}. 
PixelCNNs are a type of deep autoregressive model that computes the likelihood of an image as a factorized product of the conditional likelihood of its sub-pixels~\cite{Oord2016}. The likelihood for each pixel is modeled based on the context of preceding sub-pixels using convolutional networks (Fig.~\ref{fig1}a-b). We examine the more recent PixelCNN++ model~\cite{Salimans2017}. Despite its ability to produce accurate reconstructions and generate realistic samples, PixelCNN++'s likelihoods are also readily biased and unreliable for outlier detection~\cite{Ren2019,Serra2019,Nalisnick2019}.

We investigate the origin of the biases in PixelCNN++ likelihoods and propose effective solutions for correcting for these biases. Our contributions are as follows:
\begin{itemize}
    \item We show that biases in PixelCNN++ likelihoods arise primarily from an over-reliance of the model on local dependencies in the image.
    \item We devise two efficient solutions based on readily computed bijective transformations of the input samples. These enable correcting for biases arising from  local dependencies and pinpointing the component of the likelihood arising from long-range dependencies.
    \item The solutions we propose are computationally inexpensive to implement. Moreover, they can be applied {\it post hoc} during evaluation time and do not require retraining the model or training multiple (background) models. 
    \item We evaluate our solutions extensively using 11 inlier (5 grayscale and 6 natural image) datasets and 15 evaluation (7 grayscale and 8 natural image) datasets and show that they match or exceed state-of-the-art outlier detection performance~\cite{Ren2019,Serra2019}.
\end{itemize}

\section{Related Work}
\label{rel_work}
While considerable past work addresses outlier detection in supervised settings~\cite{Lakshminarayanan2017,liang2018}, we focus here exclusively on the unsupervised setting where no labels are available. Unlike previous deep, one-class classification approaches~\cite{andrews2016,ruff2018}, we do not employ class label information either for training or validation.

Perhaps the most relevant approach for our study is the Input Complexity (IC) metric of \cite{Serra2019}. This study characterized an identity relationship between the negative log-likelihood and image complexity, revealing a major source of bias. An elegant outlier detection score was then formulated by simply subtracting image complexity from the negative log-likelihood, with complexity being quantified as the compression length based on standard compressors (e.g., JPEG, PNG, or FLIF). In a later section, we analyze the IC metric and showcase key failure cases that violate the assumptions underlying this approach.

A second, relevant study~\cite{Ren2019} showed that a greater number of zeros in test image backgrounds biased PixelCNN++ likelihoods toward higher values. This study proposed training an additional background model with noise-corrupted images to compute a likelihood ratio that factored out the contribution of background information to the likelihood. In addition to being computationally expensive, due to the requirement of training multiple models, this metric did not perform well with our (simpler) PixelCNN++ model architecture, as we show subsequently.

A third, related study~\cite{bergman2020}  proposed an open set detection method, GOAD, that computes an anomaly score based on random affine transformations. Yet, unlike GOAD our methods involve only bijective transformations that isolate long-range dependencies in the likelihood ratio (see Sections.~\ref{sec:long_range_dep} and~\ref{sec:bijective_transformations}). Moreover, GOAD requires the transformations to be applied both during training and evaluation time whereas our approach requires transformations to be applied during evaluation time alone. Our method can, therefore, be applied to likelihoods generated by pre-trained generative models also. 

Similarly, other approaches involving generative ensembles (e.g., WAIC, ~\cite{Choi2018}) or principled statistical tests (e.g., typicality, ~\cite{Nalisnick2019b}) are either computationally expensive or do not perform well with singleton test samples, unlike our approach.

 A few studies have examined outlier detection with other classes of generative models like variational autoencoders (VAEs) or flow models. For example, \cite{Xiao2020} developed a ``Likelihood Regret'' score for robust outlier detection with VAEs. Yet, this score is expensive to compute because an optimization must be performed by retraining the VAE's encoder for each sample at test time. Similarly, \cite{Chauhan_2022} developed an efficient correction for biases in VAE visible distributions (e.g., Bernoulli) for  robust outlier detection. Both of these approaches cannot be readily extended to PixelCNN++ models. Moreover, \cite{Krichenko2020} showed that simple modifications to the architecture of normalizing flows could enable learning semantic features, thereby ameliorating low-level biases. Yet, their modification is specific to flow models and involves fully retraining the modified model. On the other hand, our approach works with a fully trained model directly at evaluation time.

Additionally, our approach resembles standard data augmentation methods, albeit superficially. For instance, \cite{yun2019} augment data by cutting and pasting image patches among training images, enabling efficient network regularization in supervised or weakly-supervised settings. Our approach, on the other hand, uses bijective transformations to isolate long-range dependencies in an unsupervised setting.

\section{De-biasing PixelCNN++ likelihoods}
\label{sec:lik_components}

\subsection{What factors contribute to the bias in PixelCNN++ likelihoods?} To analyze the bias in PixelCNN++ likelihoods, we first analyzed the factors contributing to it. To model pixel likelihoods, PixelCNN++  employs a categorical distribution approximated by a discretized mixture of logistics \cite{Salimans2017}. Variations in logistic likelihoods can be readily attributed to two sources. One source involves the location parameters (modes) of the underlying mixture of logistics: the model's best guess of the target pixel value. Accurate prediction of the target pixel value yields a higher log-likelihood (Fig.~\ref{fig1}c, left). A second source involves the scale parameters (variances): the model's uncertainty with predicting the target pixel. A lower scale parameter (greater certainty) for a mode coinciding with the target pixel value yields higher log-likelihoods (Fig.~\ref{fig1}c, right). Thus, more accurate predictions (lower reconstruction error) and more confident (less uncertain) predictions of the correct target pixel value both contribute to higher PixelCNN++ likelihoods.

We analyzed these two factors for a PixelCNN++ model trained with CIFAR10 images (ID) and tested with SVHN images (OOD) (see Fig.~\ref{fig1}d for reconstructions). The PixelCNN++ model trained on CIFAR-10 (ID) images, surprisingly reconstructs SVHN images (OOD) both with lower error (Fig.~\ref{fig1}d and~\ref{fig1}e, blue/SVHN vs. orange/CIFAR-10) and with higher confidence (lower average scale, Fig.~\ref{fig1}f, blue/SVHN vs. orange/CIFAR-10). Paradoxically, these two factors yielded ``higher likelihoods'' for SVHN/OOD than for CIFAR-10/ID samples (Fig.~\ref{fig1}g-h). Similar results were also observed with grayscale data (FMNIST/ID vs MNIST/OOD). We investigated the reasons behind these trends.

\subsection{Can global complexity fully explain the bias in PixelCNN++ likelihoods?}
\label{sec:ic}

Are biases in PixelCNN++ likelihoods fully explained by differences in overall image complexity~\cite{Serra2019}? Pixel values in less complex images are easier to model because of the stronger spatial correlations among adjacent pixels. Therefore, the comparatively lower complexity of images in datasets like MNIST or SVHN may permit more accurate and more confident predictions of sub-pixel values in these datasets, thereby inflating their likelihoods. Based on this logic, \cite{Serra2019} showed that PixelCNN++ negative log likelihoods exhibit a near-identity relationship with image complexity. They proposed an elegant OOD score that involved simply subtracting the complexity estimate ($L(\mathbf{x})$) from the negative log-likelihood (NLL) to account for this ``complexity bias''.

\begin{figure}[!t]
\centering
\includegraphics[width=0.9\linewidth]{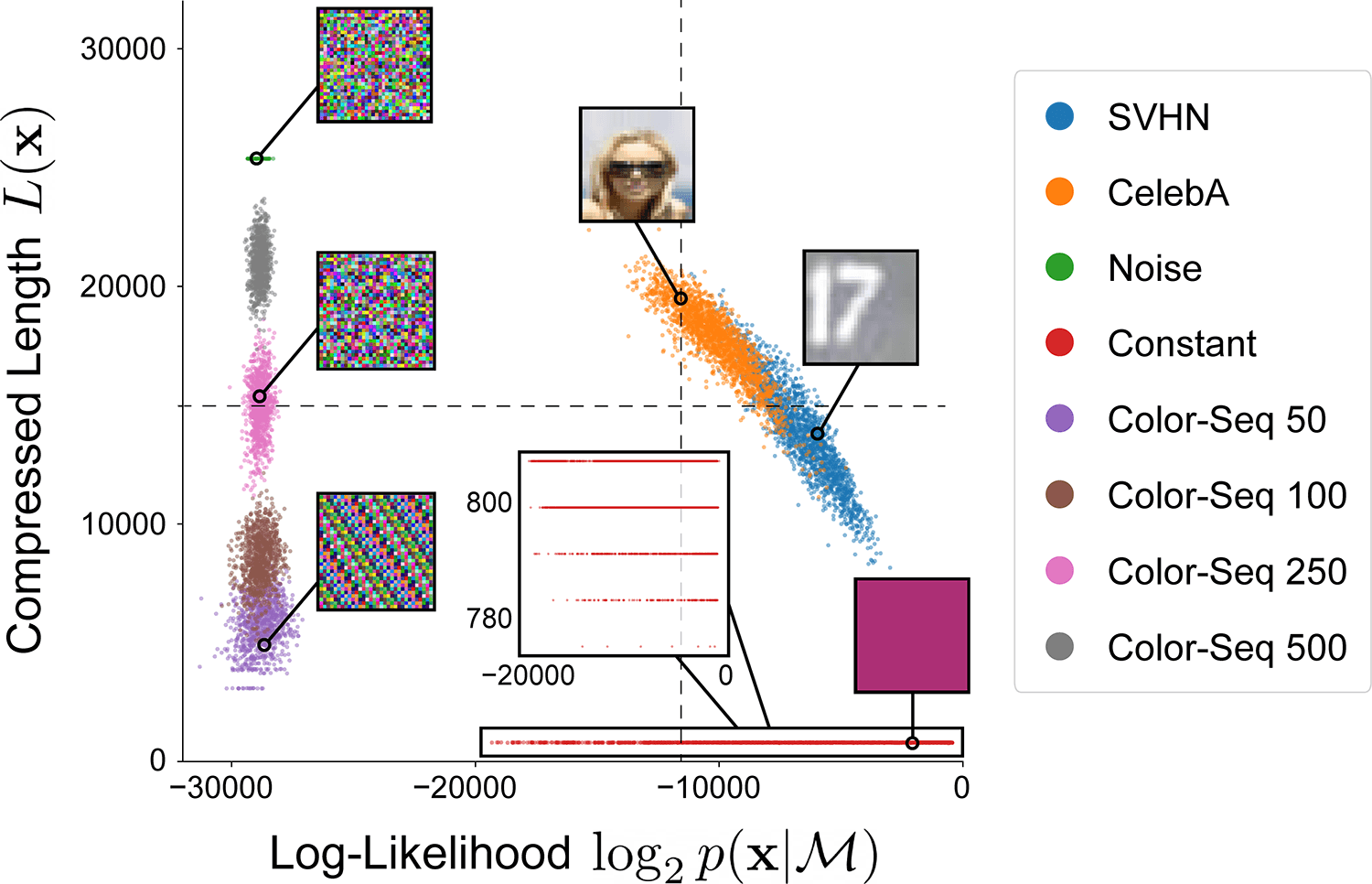}
\caption{{\bf Compressed lengths have a many-to-one relationship with PixelCNN++ likelihood.} Compressed length, measured with a PNG compressor, plotted against log-likelihood  generated by a PixelCNN++ trained on CIFAR-10 for different OOD datasets (different colors, see text). A wide range of compressed lengths occurs for the same model likelihood (dashed vertical line) and vice-versa (dashed vertical line).}
\label{fig2} 
\vspace{-1em}
\end{figure}

To further explore this assumption, we plotted normalized compressed lengths ($L(\mathbf{x})$) using a PNG compressor against log-likelihoods ($\log p(\mathbf{x})$) computed with a PixelCNN++ model trained with the CIFAR10 dataset. Although we observed a strong negative correlation between compressed lengths and log-likelihoods for OOD data comprised of natural images, this relationship was violated for other kinds of data. Specifically, ``constant'' images -- in which all pixels were of a uniform color -- revealed a nearly flat relationship (Fig.~\ref{fig2}, red points): compressed lengths were virtually identical even as log-likelihoods varied over several orders of magnitude ($\log p(\mathbf{x})$ = $\sim$-20000 to 0) depending on the image color (Fig.~\ref{fig2}, inset). By contrast, images with repeating color sequence patterns across pixels showed the opposite trend: compressed lengths varied over two orders of magnitude ($L(\mathbf{x})$ = $\sim$250 to 25000) without a substantial change in the log-likelihood (Fig.~\ref{fig2}, purple, brown, pink, and gray points). In other words, widely different log-likelihoods occurred for images with identical compressed lengths (Fig.~\ref{fig2}, dashed horizontal line). Conversely, images with similar log-likelihoods exhibited widely different compressed lengths (Fig.~\ref{fig2}, dashed vertical line).

Similar results were observed with other types of compressors (e.g., JPEG, FLIF) and also when considering the ``best'' compressor (minimum compression length, $L(\mathbf{x})$= min($L_1(\mathbf{x})$, $L_2(\mathbf{x})$, \ldots)). In other words, the assumption on which the IC metric is predicated -- that the compression length provides a reliable estimate of the negative log-likelihood under an unbiased, universal model -- appears to not hold true across all types of images. 

\begin{figure}[!ht]
\centering
\includegraphics[width=0.85\linewidth]{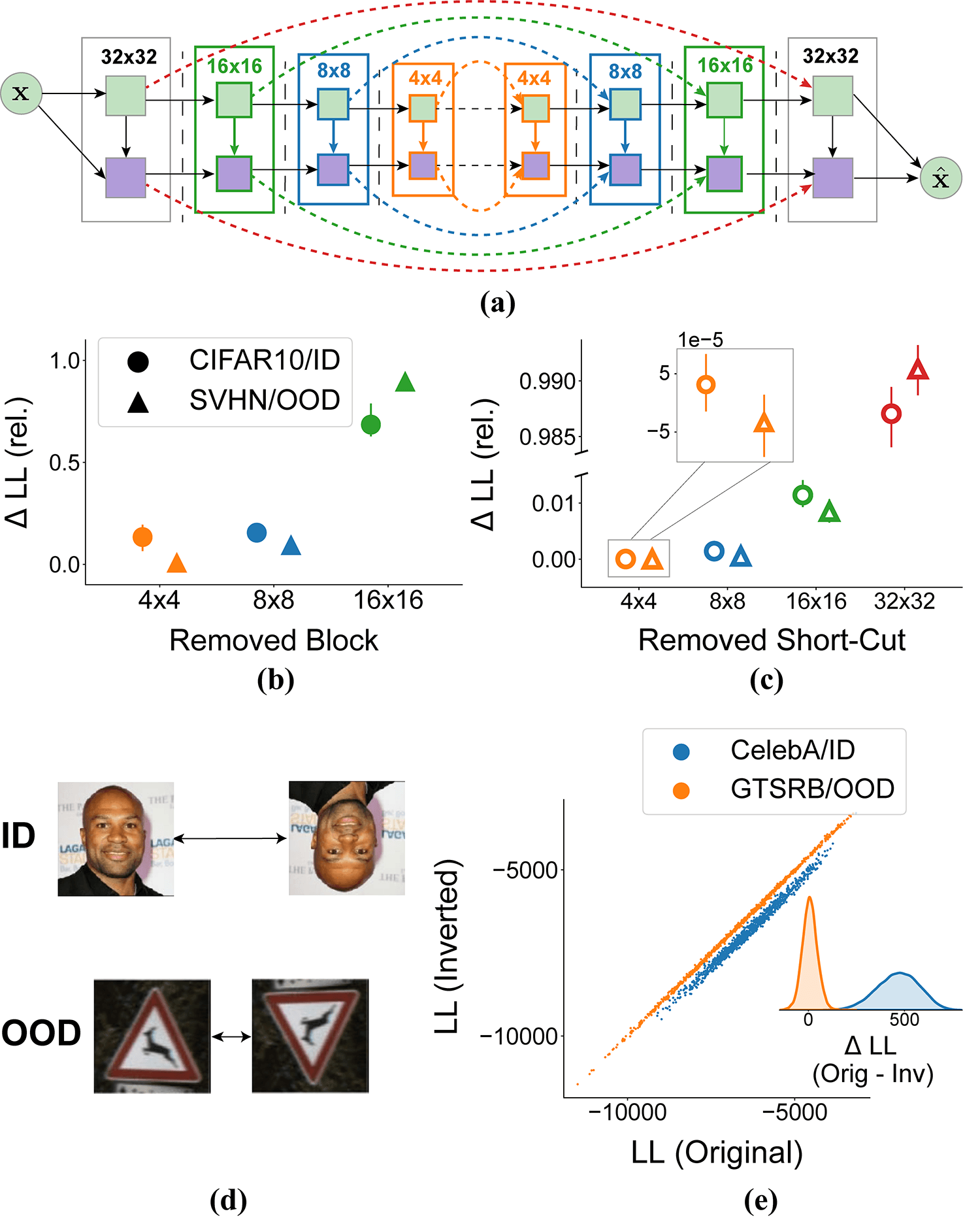}
\caption{\textbf{PixelCNN++ relies heavily on local dependencies for prediction. (a)} A PixelCNN++ model was trained on CIFAR10/ID and tested on SVHN/OOD. Ablation experiments were performed by removing the outermost (local dependency) or innermost  (long-range dependency) blocks. Blocks of matching sizes (e.g., $8 \times 8$) were ablated jointly (matching colors). In turn, additional ablations were performed by removing each set of short-cut connections. \textbf{(b)} Effect of ablating convolutional blocks on model likelihood. Orange, blue, and green: contribution of the 4$\times$4, 8$\times$8, and 16$\times$16 blocks, respectively. \textbf{(c)} Effect of ablating short-cut connections on  model likelihood. Orange, blue, green, and red: contribution of the 4$\times$4, 8$\times$8, 16$\times$16, and 32$\times$32 short-cut connections, respectively. \textbf{(d)}. A PixelCNN++ model was trained on CelebA/ID and tested on GTSRB/OOD. We expect log-likelihoods of original and inverted images to be more similar for GTSRB/OOD samples than for CelebA/ID samples. \textbf{(e)} Log-likelihoods of the inverted images plotted against their original counterparts for CelebA/ID and GTSRB/OOD. (Inset) Difference between log-likelihoods of original and inverted samples ($\Delta$LL) is higher for CelebA/ID  than for GTSRB/OOD.}
\label{fig3}
\vspace{-1em}
\end{figure}

\subsection{Do local dependencies contribute to the bias in PixelCNN++ likelihoods?}
\label{sec:local_dep}
In addition to global complexity, could local dependencies across pixels bias PixelCNN++ likelihoods? We explored the hypothesis that PixelCNN++ leverages local dependencies to generate accurate and confident predictions. For example, both SVHN and MNIST images typically comprise digits with relatively simple features embedded in a fairly uniform background. By acquiring knowledge of simple local features like edges and contours, the model can accurately predict sub-pixel values in adjacent pixels. In fact, this could happen even for OOD images, using knowledge of the local neighborhood, without learning the long-range structure of the data.

We tested this hypothesis using simple ablations to a standard PixelCNN++  model (Fig.~\ref{fig3}a). In the model, the innermost parts of the network capture long-range dependencies over longer spatial scales -- a direct consequence of using strided convolutions over progressive layers~\cite{Salimans2017}. 
In our PixelCNN++ model, in addition to the 8$\times$8 and 16$\times$16 CNN layers in the standard model, we introduced another sequence of 4$\times$4 layers in the innermost part of the network (Fig.~\ref{fig3}a, orange block).

We tested the effect of progressively removing nested hierarchies of the innermost 4$\times$4, 8$\times$8, and 16$\times$16 convolutional layers on model likelihoods. 
Precisely in line with our hypothesis, removing the 4$\times$4 layers (Fig.~\ref{fig3}b, orange triangle) produced virtually no change in the log-likelihoods (LLs) for OOD data (SVHN OOD vs CIFAR-10/ID). Yet, ablating the 8$\times$8 (Fig.~\ref{fig3}b, blue triangle) or 16$\times$16 (Fig.~\ref{fig3}b, green triangle) innermost blocks produced substantially greater reductions in likelihoods. In other words, the model relied primarily on local dependencies when making predictions with OOD data. By contrast, for ID test data, even just removing the 4$\times$4  block (Fig.~\ref{fig3}b, orange circle) produced a noticeable change in the likelihoods, indicating that the model had learned to exploit long-range dependencies when making predictions with ID data. These progressive changes in likelihoods were accompanied by progressively higher reconstruction errors and more uncertain predictions in the model. Removing the short-cut connections, at different levels, in turn, yielded similar results, with the largest reduction in likelihoods occurring when the outermost layer (32$\times$32) connections (Fig.~\ref{fig3}c, red symbols) were removed. As before, ablating the innermost layers had a more significant effect on ID likelihoods than OOD data (Fig.~\ref{fig3}c, circles vs triangles). 

In sum, the PixelCNN++ model relied strongly on local dependencies for accurate and confident predictions. Yet, removing network components that captured long-range dependencies (innermost layers) produced larger changes in likelihoods for ID than for OOD samples. We sought to exploit these differences for efficient outlier detection.

\begin{figure*}[tbh]
\centering
\includegraphics[width=0.8\linewidth]{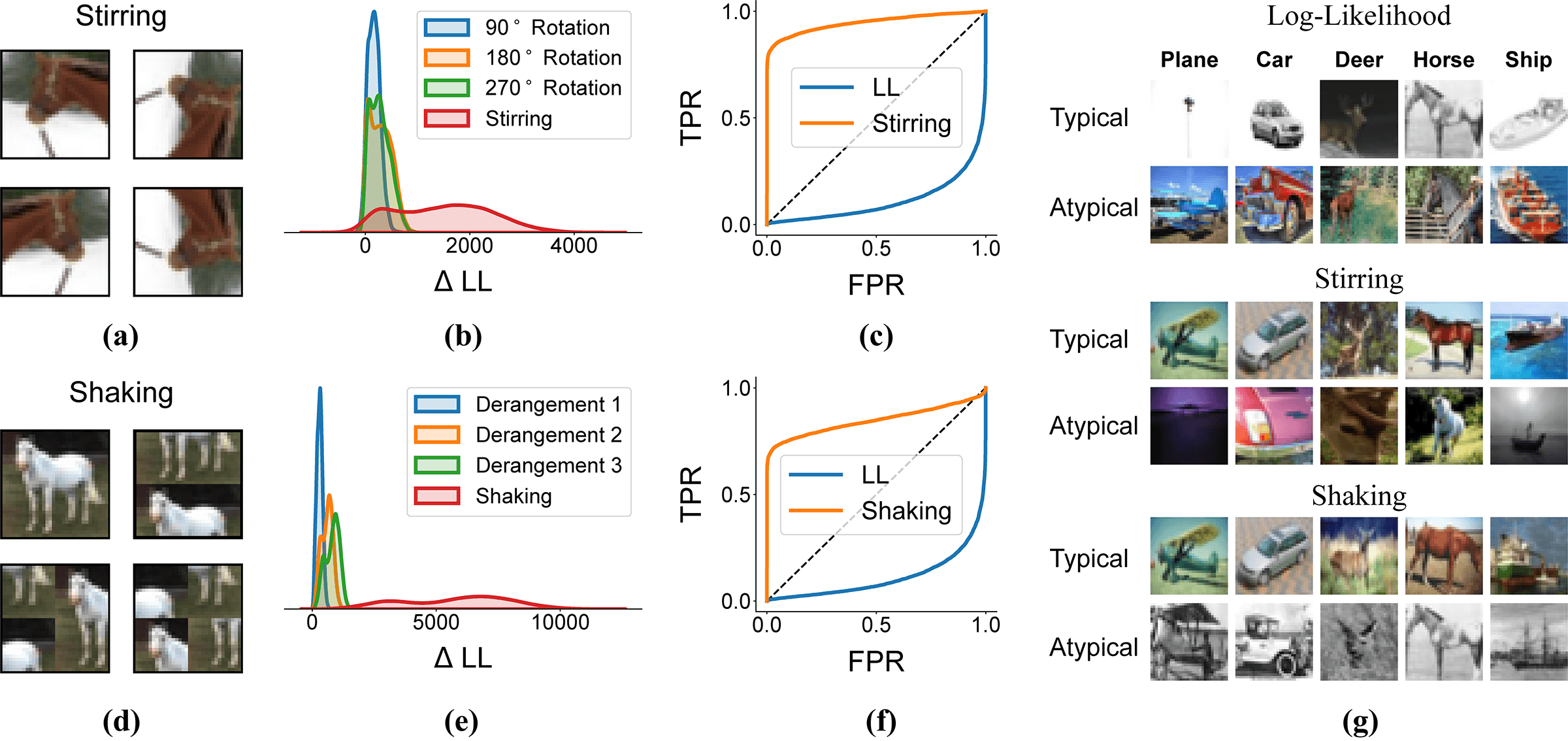}
\caption{{\bf Robust outlier detection with ``stirring'' and ``shaking''.} \textbf{(a)} Examples of ``stirring'' transformations. \textbf{(b)} Change in PixelCNN++ LL after ``stirring'', (red) as compared with rotations,  applied individually (CIFAR10/ID). \textbf{(c)} ROC curve for outlier detection for CIFAR10/ID vs SVHN/OOD case using vanilla log-likelihood (blue) and ``stirring'' (orange). \textbf{(d)} Examples of ``shaking'' transformations \textbf{(e, f)} Same as in (b) and (c), but for ``shaking''. \textbf{(g)} Specific sub-category of CIFAR10/ID images (columns) that were assigned among the highest and lowest vanilla log-likelihoods  $\log p(\mathbf{x})$ (top 2 rows), following ``stirring'' (middle 2 rows) or following ``shaking'' (bottom 2 rows).}
\label{fig5}
\vspace{-1em}
\end{figure*}

\subsection{Isolating the contribution of long-range dependencies to PixelCNN++ likelihoods}
\label{sec:long_range_dep}

Given the strong bias in PixelCNN++ likelihoods induced by local dependencies, we asked whether removing contributions of local dependencies or isolating contributions of long-range dependencies would de-bias PixelCNN++ likelihoods. To this end, we explored simple transformations to the input images that would preserve local dependencies but systematically perturb long-range dependencies. We hypothesized that perturbing long-range (but not local) dependencies would produce a stronger degradation of the likelihood for ID data as compared to OOD data. 

To illustrate this idea, we explore a transformation by ``inversion''. Because the PixelCNN++ model has a specific order of scanning and predicting pixels in the image (upper left to lower right), upon inversion, characteristic image features fall into a context unfamiliar to the model. For example, inverting an image of a face positions the eyes below the nose, and the nose below the mouth (Fig.~\ref{fig3}d, top). A PixelCNN++ model trained, for example, on a dataset with face images (e.g., CelebA) would then predict pixels in an inverted face image less accurately and with higher uncertainty than pixels in a standard, upright face. As a result, the model would yield lower likelihoods for inverted CelebA images rather than upright images. We hypothesized that this would not occur for OOD images (e.g., GTSRB). In this case, the model is unlikely to rely on long-range predictions for either the upright or the inverted GTSRB images because both categories of images are equally unfamiliar  (Fig.~\ref{fig3}d, bottom). Thus, the model should yield equivalent likelihoods for both inverted and upright GTSRB images.

We tested and confirmed this hypothesis with the PixelCNN++ model trained on the CelebA dataset. The model yielded systematically lower likelihoods for inverted faces than for upright faces in the ID data (Fig.~\ref{fig3}e, blue points). In contrast, the model yielded virtually identical likelihoods for OOD (GTSRB) images (Fig.~\ref{fig3}e, orange points). Therefore, one solution for outlier detection is to simply subtract the log-likelihood of the original image from that of the perturbed (inverted) image to isolate the contribution arising from long-range dependencies. With this reasoning, we propose an ``outlier detection score'' as follows:

$$
\log p_{\mathrm{LR}}(\mathbf{x}) = \log p_\theta(\mathbf{x}) - \log p_\theta(\mathbf{x'})
$$

where $\theta$ represents the PixelCNN++ model parameters, $\mathbf{x}$ represents the test sample (image, in this case), $\mathbf{x'}$ represents the same test sample after a perturbation that preserves local dependencies but disrupts long-range dependencies, $\log p_\theta(\mathbf{x})$ represents the log-likelihood of sample yielded by the PixelCNN++ model and $\log p_{\mathrm{LR}}$ represents a component of the log-likelihood that depends primarily on long-range dependencies in the model. This formulation can also be construed as a log-likelihood ratio between the original and perturbed samples, assuming factorizable contributions of local and long-range dependencies to the overall likelihood. We expect to observe a larger $\log p_{\mathrm{LR}}$ for ID data than OOD data.

Central to the success of this approach is identifying transformations that preserve local dependencies while disrupting long-range ones. We identify and explore two families of  transformations that we call ``stirring'' and ``shaking''.

\subsection{Bijective transformations for robust outlier detection with PixelCNN++}
\label{sec:bijective_transformations}

{\bf Stirring.} We extend the inversion solution by incorporating a family of 7 geometric transformations, including 3 rotations of the original image (by 90$^{\circ}$, 180$^{\circ}$ or 270$^{\circ}$), lateral inversion (mirror reflection about the vertical midline), and 3 rotations of the reflected image (again, by 90$^{\circ}$, 180$^{\circ}$ or 270$^{\circ}$). The $\log p_{\mathrm{LR}}$ is summed across all 7 transformations to yield the final outlier detection score. We term the collection of these transformations as ``stirring'' (Fig.~\ref{fig5}a). Because the individual transformations contribute additively, ``stirring'' produced a larger change in likelihoods for the perturbed images compared to the upright images (Fig.~\ref{fig5}b, red density) than individual rotations (Fig.~\ref{fig5}b, blue, orange, and green densities). ``Stirring'' may, thus, enable robust outlier detection for  images with distinct axes of symmetry.

{\bf Shaking.} We consider a second class of bijective transformations that involve dividing the images into patches and  shuffling these patches randomly.
We consider three ways to achieve this: i) splitting the image in half along the horizontal midline, ii) splitting the image in half along the vertical midline, and iii) splitting the images into four quarters along the horizontal and vertical midlines (Fig.~\ref{fig5}d). These permit a total of 9 unique derangements -- random permutations in which no patch is located in its original position. $\log p_{\mathrm{LR}}$ is summed across all 9 derangements to yield the final outlier detection score. We term the collection of these derangements as ``shaking'' (Fig~\ref{fig5}d). Again, ``shaking'' produced a larger change in likelihoods for the perturbed images compared to the upright images than individual derangements (Fig.~\ref{fig5}e).

\begin{figure*}[thb]
\centering
\includegraphics[width=0.9\textwidth]{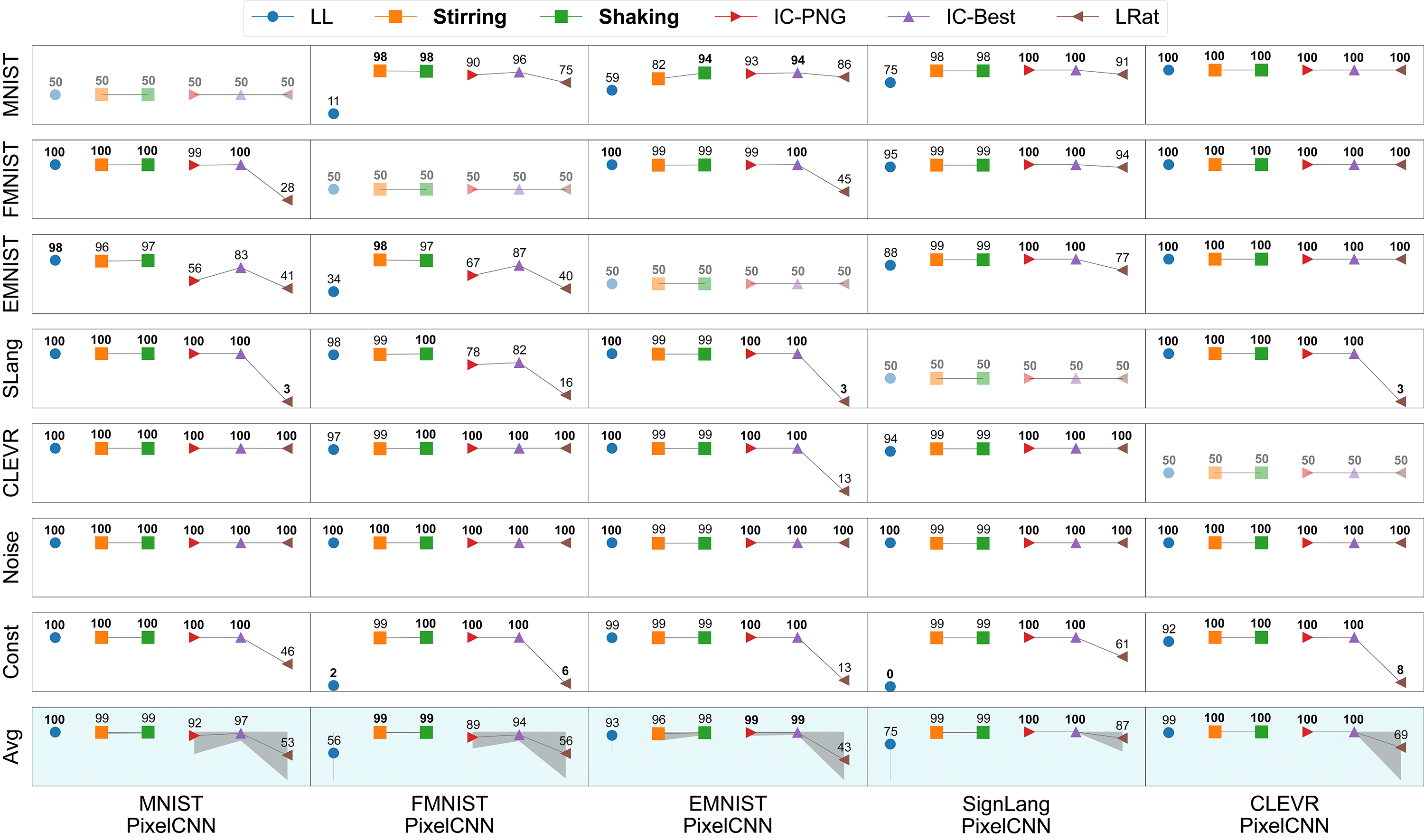}
\caption{\textbf{Outlier detection performance: Grayscale data}. Outlier detection AUROC values for PixelCNNs trained with grayscale image datasets (ID, columns) and tested with other grayscale datasets (OOD, rows). Last row: average AUROC; gray shading: range of AUROC values. Blue: log-likelihood (LL), uncorrected; Orange: ``Stirred'' LL; Green: ``Shaken'' LL; Red: Input Complexity (PNG); Purple: Input Complexity (Best); Brown: Likelihood Ratio. Numbers in bold: best performance.}
\label{fig:grayscale}
\vspace{-1em}
\end{figure*}

{\bf Conditional correction}. Our outlier detection score is a ratio of the logarithm of two probability densities: $p_\theta(\mathbf{x})$ and $p_\theta(\mathbf{x'})$. When $p_\theta(\mathbf{x})$ is a very small numerical value, the likelihood of the perturbed sample $\mathbf{x'}$, $p_\theta(\mathbf{x'})$ would also be comparably small numerically. In this case, it is likely, that the estimation of $\log p_{\mathrm{LR}}$ would be noisy and far from accurate. To avoid such noisy estimates, we adopt a pre-filtering strategy based on identifying outliers with the model log-likelihood alone, following which ``stirring'' and ``shaking'' corrections are applied. We perform ablation experiments to estimate the contribution of this conditional correction.

\section{Experiments}
We trained PixelCNN++ models on each of five grayscale image datasets: MNIST, FashionMNIST, EMNIST Letters, Sign Language MNIST, and CLEVR \cite{mnist,fmnist,emnist,clevr,signlang}. Each of these models was tested against six OOD datasets, including the other four datasets and noise and constant images. Similarly, we trained PixelCNN++ models on each of six  natural image datasets - SVHN, CelebA, CompCars, GTSRB, CIFAR10, and LSUN (classroom) \cite{svhn,celeba,compcars,gtsrb,cifar10,lsun}. Each model was tested against seven OOD datasets, including noise and constant images. 

We report the area under the ROC curve (AUROC) between the respective test sets of the ID and OOD datasets in a 7$\times$5 grid for grayscale and an 8$\times$6 grid for natural image data. All results reported include the conditional correction.

\begin{figure*}[thb]
\centering
\includegraphics[width=0.95\textwidth]{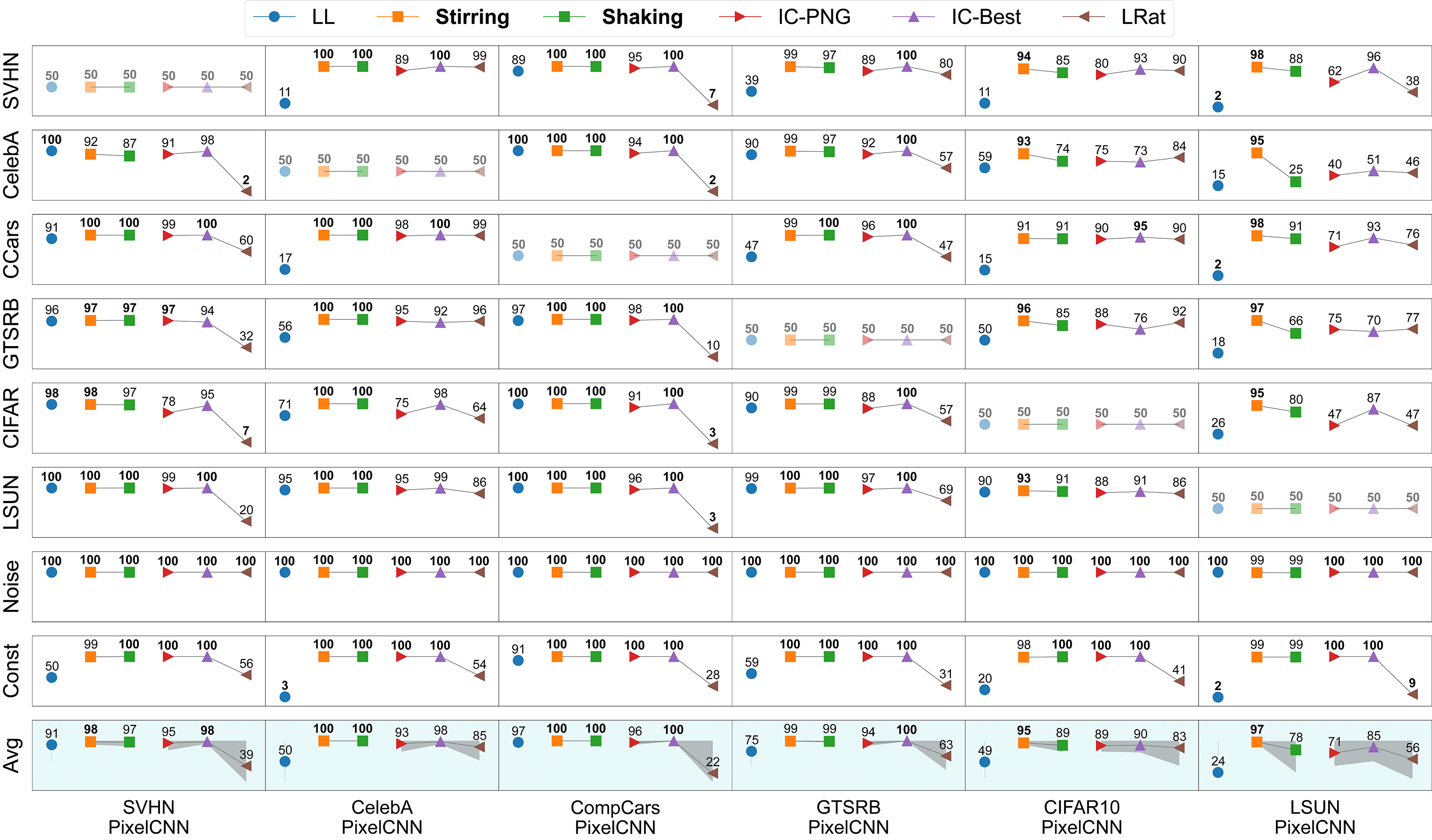}
\caption{\textbf{Outlier detection performance: Natural image data.} Same as in Figure~\ref{fig:grayscale} but for natural image datasets.}
\label{fig:color}
\vspace{-1em}
\end{figure*}

\subsection{Outlier detection performance with ``stirring'' and ``shaking''}
For illustration, we compare outlier detection performance of $\log p_{\mathrm{LR}}$ with ``stirring'' with the vanilla log-likelihood. We obtained state-of-the-art AUROCs ($\mathtt{\sim}$95\%) for the particularly problematic case of CIFAR10 ID versus SVHN OOD (Fig.~\ref{fig5}c). Moreover, typical exemplars were assigned among the highest $\log p_{\mathrm{LR}}$ (Fig.~\ref{fig5}g, middle), whereas this was not the case using the vanilla log-likelihoods (Fig.~\ref{fig5}g, top).

This superlative performance was observed across all other comparisons comprising both grayscale and natural image datasets.  In almost all cases, ``stirring'' (Fig.~\ref{fig:grayscale} and Fig.~\ref{fig:color}, orange symbols) outperformed vanilla LL (Fig.~\ref{fig:grayscale} and Fig.~\ref{fig:color}, blue symbols). ``Stirring'' achieved a performance near-ceiling in most of the grayscale cases. In the challenging cases of FMNIST/ID, CIFAR10/ID, and LSUN/ID,  ``stirring'' achieved AUROCs of 95 and above. Overall, with ``stirring'', we saw an average AUROC improvement of $\mathtt{\sim}$14\% for grayscale images and $\mathtt{\sim}$57\%  for natural images.

 
We also obtained similar results with $\log p_{\mathrm{LR}}$ computed from ``shaken'' images. Again, ``shaking'' improved outlier detection for the challenging case of CIFAR10/ID versus SVHN/OOD (Fig.~\ref{fig5}f). In general, outlier detection with ``shaking'' improved  across both grayscale and natural images (Fig.~\ref{fig:grayscale} and Fig.~\ref{fig:color}, green symbols), as compared to that with vanilla log-likelihoods. Overall, outlier detection performance improved, on average, by $\mathtt{\sim}$14\% for grayscale images and $\mathtt{\sim}$47\% for natural images, indicating marginally worse improvements than with ``shaking''. Interestingly, with ``stirring'', color information played a major role in determining atypical exemplars (Fig.~\ref{fig5}g, lower).

\subsection{Comparison with competing methods}

We compare our results with two state-of-the-art methods -- Likelihood Ratios \cite{Ren2019}, and Input Complexity \cite{Serra2019} -- the two most relevant competing approaches for state-of-the-art outlier detection with PixelCNN++ (see section \ref{rel_work}).

Our methods comfortably outperformed likelihood ratios (Fig.~\ref{fig:grayscale} and Fig.~\ref{fig:color}, filled red symbols) in all cases. ``Stirring'' performed $\mathtt{\sim}$48\% better on average for grayscale images and $\mathtt{\sim}$80\% better for natural images than likelihood ratios. Similarly, ``shaking'' performed $\mathtt{\sim}$49\% better, on average, for grayscale images and $\mathtt{\sim}$69\% better for natural images than likelihood ratios. The OOD detection numbers that we report for likelihood ratios are poorer than those reported by ~\cite{Ren2019}, who used a more complex model architecture; these results suggest that the success of the likelihood ratio metric is architecture dependent. 

Our metrics also outperformed or performed comparably with Input Complexity computed using the PNG compressor (Fig.~\ref{fig:grayscale} and Fig.~\ref{fig:color}; IC-PNG, brown symbols) or using the Best compressor (minimum compressed length; IC-Best, purple symbols). ``Stirring'' performed $\mathtt{\sim}$6\% (2\%) better, on average, for grayscale images and $\mathtt{\sim}$14\% (5\%) better for natural images than IC-PNG (IC-Best). Similarly, ``shaking'' performed $\mathtt{\sim}$7\% (2\%) better, on average, for grayscale images and $\mathtt{\sim}$7\% (-2\%) better for natural images than IC-PNG (IC-Best). Surprisingly, IC-Best did not perform well with specific datasets, like LSUN/ID or CIFAR10/ID, on which our metrics, especially those based on ``stirring'', performed exceedingly well. 

\subsection{Time and Space Complexity}
\label{time_space_complexity}

We also compared the time and space complexity of our methods with competing methods. We measured time complexity as the average per sample inference time for computing the OOD detection score. We quantified space complexity as the peak memory usage (mebibytes/MiB) during inference. Both metrics were computed with all the natural image datasets. Table.~\ref{tab:time_space} shows these metrics for ``stirring'' and ``shaking'' alongside those for competing methods. Our methods do not require training additional background models and are competitive with state-of-the-art in terms of time and space complexity.

\begin{table}[h!]
\centering
\begin{tabular}{lrr}
\toprule
Method & Time / Sample (ms) & Peak Mem Usage (MiB)\\
\midrule
Stirring & 14.07 & 3142.76\\
Shaking & 18.02 & 3289.24\\
LRat & 7.95 + 2580.10* & 3634.00\\
IC & 4.69 & 3120.80\\
\bottomrule
\multicolumn{3}{r}{\small (*additional training time for background model)}\\
\end{tabular}
\caption{\textbf{Time and Space Complexity.} Time complexity (inference time per sample) and space complexity (peak memory usage during inference) for ``stirring'', ``shaking'', likelihood ratios (LRat) \protect\cite{Ren2019}, and Input Complexity (IC) \protect\cite{Serra2019}.}
\label{tab:time_space}
\end{table}

\subsection{Results with other generative models}
\label{sec:other_gen_models}

\begin{figure*}[hbt]
    \centering
    \begin{subfigure}[b]{0.49\textwidth}
        \centering
        \includegraphics[width=0.95\linewidth]{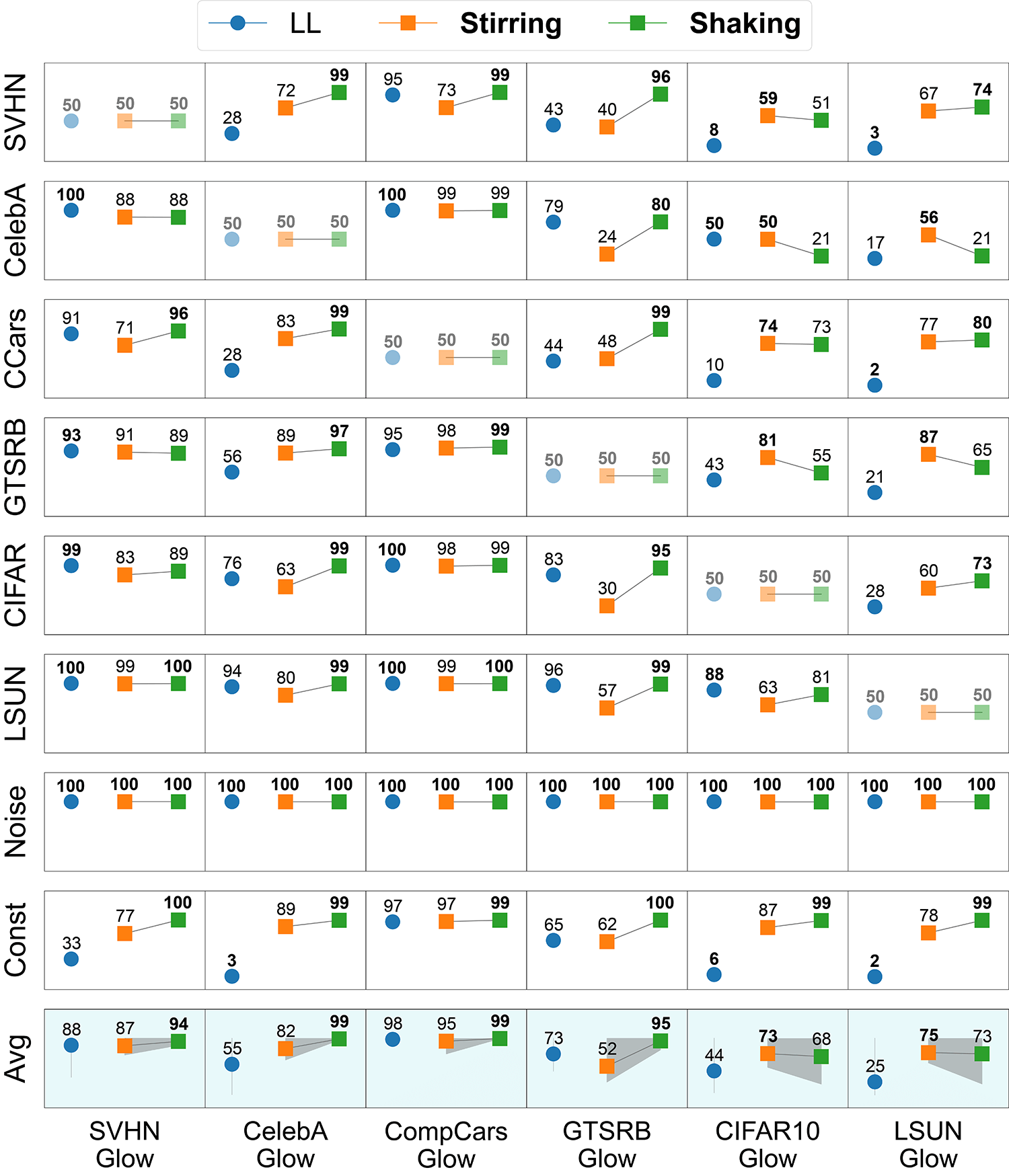}
        \caption{}
        \label{fig:glow_color}
    \end{subfigure}
    \begin{subfigure}[b]{0.49\textwidth}
        \centering
        \includegraphics[width=0.95\linewidth]{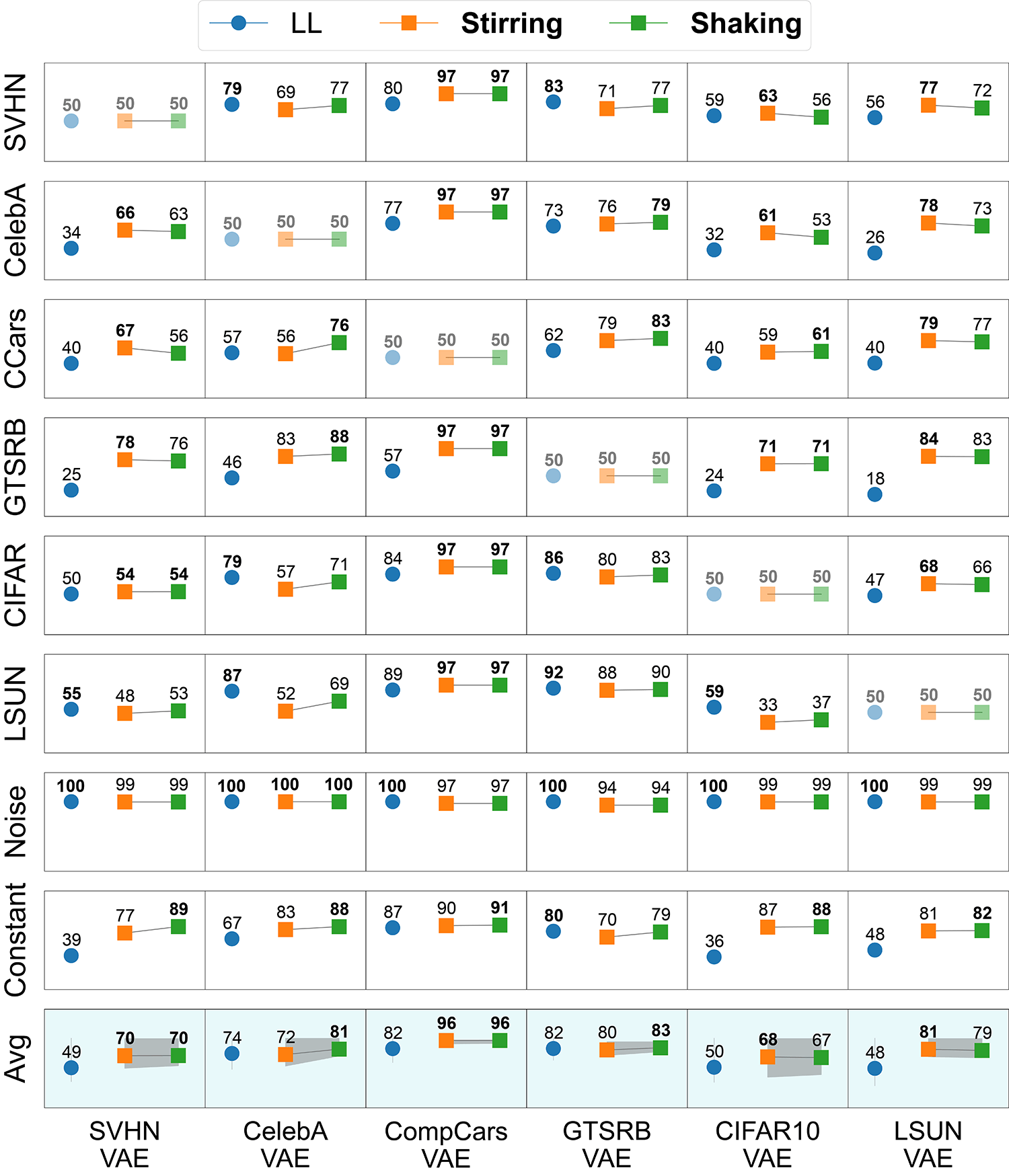}
        \caption{}
        \label{fig:vae_color}
    \end{subfigure}
    \caption{\textbf{Outlier detection performance (AUROC) with Glow and VAE.} \textbf{(a)} Outlier detection AUROC values for Glow models trained on natural image datasets. Conventions are the same as in Figure.~\ref{fig:color}.  \textbf{(b)} Outlier detection AUROC values for VAEs trained on natural image datasets. Conventions are the same as in Figure.~\ref{fig:color}.}
    \label{fig:other_gen_models}
    \vspace{-1em}
\end{figure*}

Generative models come in at least three major flavors~\cite{glow}: i) Autoregressive models, ii) Variational Autoencoders, and iii) Flow-based models.

While we have shown that our approaches work well with PixelCNN++, a model of the first category, we tested whether these approaches work with the other two models also: Generative flows \cite{glow} and Variational Autoencoders \cite{Kingma2013}.

\subsubsection{Generative Flows}
\label{app:glow_results}

Flow-based models are a popular class of DGMs, as they enable computing exact log-likelihoods; this makes it a ready choice for OOD detection tasks. Yet, these models were also shown to assign higher likelihoods to OOD than ID images \cite{Nalisnick2019}. 

We quantified OOD detection performance with natural image datasets using Glow log-likelihoods. These vanilla log-likelihoods  (Fig.~\ref{fig:glow_color}, blue symbols) failed in several OOD detection cases. Like with PixelCNN++, the clearest failure cases occurred with CIFAR10/ID and LSUN/ID.

We then applied ``shaking'' and ``stirring'' to the Glow model likelihoods. In addition, we employed the conditional correction as specified in section~\ref{sec:bijective_transformations} except for the following modification: because the training log-likelihoods were not normally distributed, we used the 99.5th percentile of the training data as a cutoff, instead of the 3-MAD criterion. 

Both ``stirring'' (Fig.~\ref{fig:glow_color}, orange symbols) and ``shaking'' (Fig.~\ref{fig:glow_color}, green symbols) generally improved OOD detection performance. ``Stirring'' performed $\mathtt{\sim}$17\% better, on average, than vanilla log-likelihoods. Similarly, ``shaking'' performed $\mathtt{\sim}$38\% better, on average, than vanilla log-likelihoods.

\subsubsection{Variational AutoEncoder}
\label{app:vae_results}

Variational Autoencoders \cite{Kingma2013} are another class of DGMs that enable estimating sample likelihoods using variational inference, which renders these relevant for OOD detection tasks. VAEs are also considered unreliable for OOD detection tasks ~\cite{Ren2019,Nalisnick2019,Choi2018,Xiao2020,Chauhan_2022}.

As before, we quantified OOD detection performance with vanilla log-likelihoods (Fig.~\ref{fig:vae_color}, blue symbols); several failure cases occurred. Again, we applied ``shaking'' and ``stirring'', as with the Glow models, except we also applied contrast stretching and a bias correction for the reconstruction error, following \cite{Chauhan_2022}. These additional steps were necessary as ``stirring'' and ``shaking'' are not designed to rectify biases in the reconstruction error arising from the VAE visible distribution (see \cite{Chauhan_2022} for details). 

With this approach, OOD detection performance improved. ``Stirring'' (Fig.~\ref{fig:vae_color}, orange symbols) performed $\mathtt{\sim}$21\% better, on average, than vanilla log-likelihoods. Similarly, ``shaking'' (Fig.~\ref{fig:vae_color}, green symbols) performed $\mathtt{\sim}$24\% better, on average, than vanilla log-likelihoods. 

\subsubsection{Analysis of differential efficacy across DGMs}

\begin{figure}[h!]
    \centering
    \includegraphics[width=0.8\linewidth]{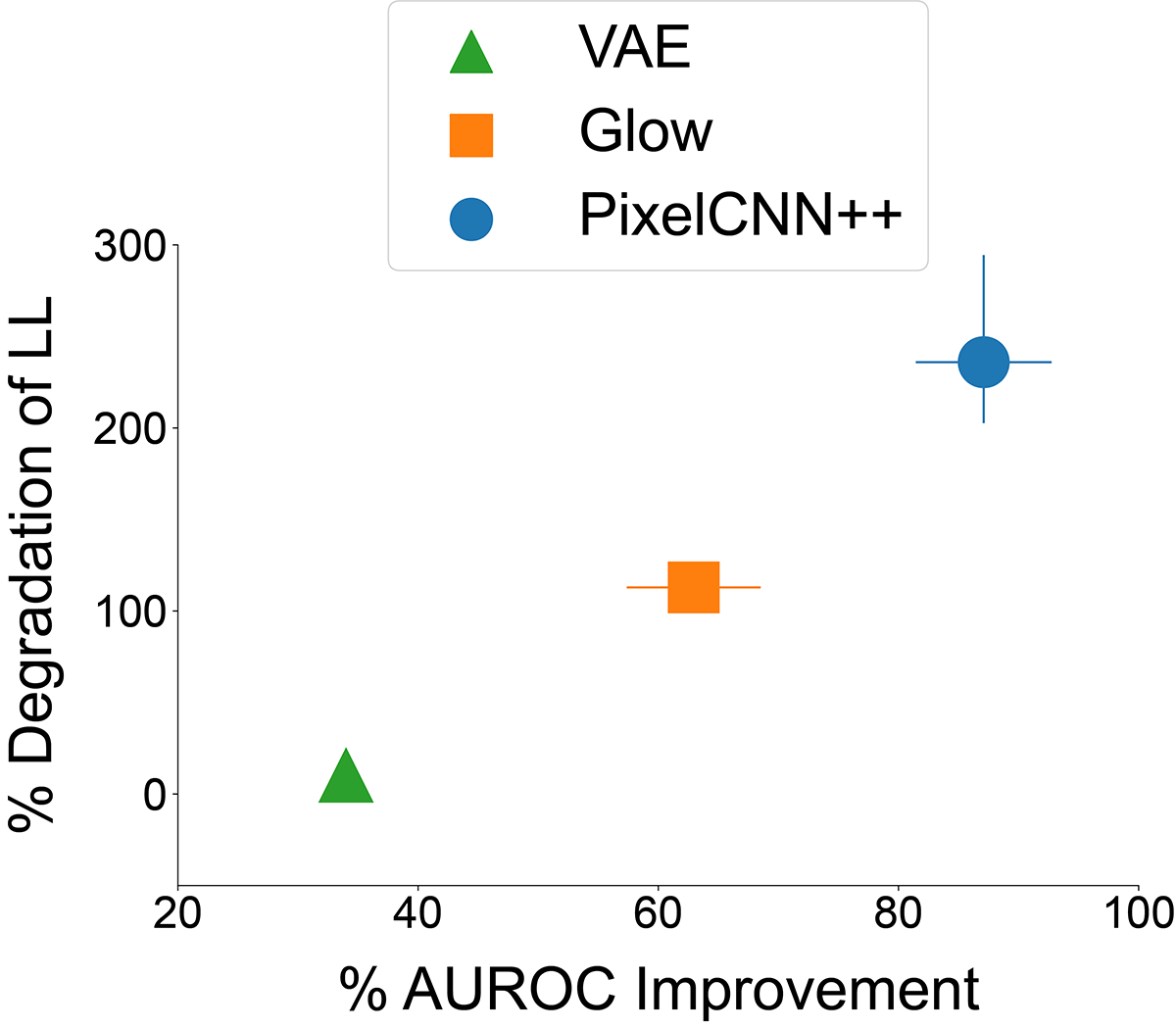}
    \caption{\textbf{Efficacy of ``stirring'' and ``shaking'' across different deep generative model classes.} Percentage improvement in AUROC (average of ``stirring'' and ``shaking'') over vanilla log-likelihood (x-axis) plotted against the \% degradation in sample log-likelihood after perturbating a local neighborhood surrounding each pixel (CIFAR10/ID; see text for details). Error bars: range of performance improvements across ``stirring'' and ``shaking'' (x-axis) and 25th and 75th percentiles of log-likelihood degradation (y-axis). In some cases, the error bars are smaller than the symbol sizes.
    }
    \label{app:fig:dgm_efficacy}
    \vspace{-1em}
\end{figure}

Comparing improvements across DGM classes, we observed that our approaches were most effective with PixelCNN++ models, but comparatively less effective with Glow, and least effective with VAEs (Fig.~\ref{app:fig:dgm_efficacy}, x-axis). We hypothesize that the efficacy of our approaches was higher for models that relied more on local dependencies for likelihood estimation. 

To test this hypothesis, we replaced a local neighborhood (3x3 patch) surrounding a given pixel, with uniform random values (0-255); the replacement was performed once per pixel for each test image for the CIFAR10/ID dataset. With this perturbation, we quantified the degradation in sample log-likelihood, as a percentage relative to its original value. The degradation was greatest for PixelCNN++, followed by Glow and VAE in that order (Fig.~\ref{app:fig:dgm_efficacy}, y-axis), reflecting the extent to which each model relied on local dependencies for predicting pixel values (VAE$<$Glow$<$PixelCNN++). In other words, the efficacy of our approaches was indeed governed by how much each model relied on local dependencies for assigning sample likelihoods.



\section{Discussion}
We developed simple, lightweight, and intuitive methods for state-of-the-art image outlier detection with PixelCNN++. Although we motivated our approaches with this particular class of deep generative model, we showed that they generalize to other classes of generative models, like Glow and VAEs, also. In addition, our methods may be relevant for outlier detection with other types of temporal data (e.g., speech), for which ``stirring'' and ``shaking'' can be readily performed, either by reversing the data in time or by chunking and shuffling the sequence.

Across all 72 ID/OOD comparisons with PixelCNN++, ``stirring'' performed noticeably worse than ceiling performance only with  EMNIST/ID versus MNIST/OOD (AUROC of $\mathtt{\sim}$82\%), a not unreasonable failure given the feature similarity of MNIST and EMNIST samples. In fact, IC exhibited confusion for these two datasets also (MNIST/ID versus EMNIST/OOD, IC-Best AUROC $\mathtt{\sim}$83\%). On the other hand, while ``shaking'' performed well with most datasets, improvements were less impressive than those with ``stirring'', especially for natural ID image datasets like CIFAR10 and LSUN. 

Our approaches could also be relevant for outlier detection with more recent autoregressive models. To our knowledge, such models (e.g., autoregressive transformers \cite{cao2021}) have not been widely used for unsupervised outlier detection. Moreover, deep diffusion models require non-standard approximations for maximum likelihood training \cite{song2021}, precluding their widespread use in OOD detection. Future work will test whether biases due to local dependencies persist in these recent models also.

More generally, our results gainsay widespread claims in recent literature -- that likelihoods from deep generative models are unreliable for outlier detection~\cite{Nalisnick2019,Ren2019,Serra2019}. Rather, PixelCNN++ models are exquisitely sensitive to long-range dependencies in their training data, and isolating these dependencies, with simple geometric transformations, suffices to achieve robust outlier detection.

\section*{Acknowledgements}
This work was conducted as part of a sponsored Google Research project. The authors wish to thank Dr. Manish Gupta for helpful discussions during the course of the study.

\bibliographystyle{named}
\bibliography{ijcai23}

\clearpage 

\appendix

\begin{center}
{\bf Shaken, and Stirred: Long-Range Dependencies Enable Robust Outlier Detection with PixelCNN++}
\end{center}

\section*{Appendices \\}

\section{Model Architecture and Training}
\label{app:architecture}

\begin{table}[h]
\caption{{\bf PixelCNN++ architecture} Parameters used in TensorFlow Probability's implementation of PixelCNN++.}
\centering
\small
\resizebox{\linewidth}{!}{%
\begin{tabular}{lrr}
\toprule
\textbf{Parameter} & \textbf{Grayscale} & \textbf{Natural}\\

\midrule
image\_shape & (32, 32, 1) & (32, 32, 3)\\
conditional\_shape * & None & None\\
num\_resnet & 2 & 2\\
num\_hierarchies & 4 & 4\\
num\_filters & 32 & 64\\
num\_logistic\_mix & 5 & 5\\
receptive\_field\_dims * & (3, 3) & (3, 3)\\
dropout\_p & 0.3 & 0.3\\
resnet\_activation * & `concat\_elu' & `concat\_elu'\\
use\_weight\_norm * & True & True\\
use\_data\_init * & True & True\\
high * & 255 & 255\\
low * & 0 & 0\\
dtype * & tf.float32 & tf.float32\\
\bottomrule
\multicolumn{3}{l}{* indicates arguments where we used TensorFlow defaults}
\end{tabular}
}

\label{tab:arch}
\end{table}

We use TensorFlow Probability's\footnote{\url{https://www.tensorflow.org/probability}} implementation of PixelCNN++. Unlike the older PixelCNN model, PixelCNN++ employs downsampling and upsampling at specific layers to capture long-range correlations. We follow the same idea and employ an architecture similar to that in the original PixelCNN++ paper. Our model consists of 8 blocks of 2 residual layers each. We use 64 (3$\times$3) convolutional filters in all layers for natural images and 32 (3$\times$3) convolutional filters for grayscale images. Between the first and the second blocks, the second and the third blocks, and the third and the fourth blocks, we perform downsampling by a factor of 2 with strided convolutions. We place a skip (identity) connection between the fourth and the fifth blocks. We perform upsampling by a factor of 2 using strided transpose convolutions between the fifth and the sixth blocks, sixth and seventh blocks, and seventh and eighth blocks. One key difference with the standard PixelCNN++ model is that our hierarchies are 4 levels deep so that our model has 32$\times$32, 16$\times$16, 8$\times$8 and 4$\times$4 size convolutional layers (Fig.~\ref{fig1}a). The inner layers were introduced in  PixelCNN++  specifically to capture long-range dependencies~\cite{Salimans2017} -- a feature absent in the original PixelCNN model~\cite{Oord2016} -- and our architecture increases the nesting of the inner layers by having one level more.

In addition, we employ short-cut connections between residual layers of matched size. In particular, blocks one and eight, two and seven, three and six, four and five are connected using short-cut dense connections (Fig.~\ref{fig1}a, curved, dashed). We perform regularization by setting the dropout parameter to 0.3. Finally, we employ a mixture of 5 logistics to approximate the categorical likelihoods. The parameters of ``tfp.distributions.PixelCNN'' required to replicate our model architecture are shown in Table.~\ref{tab:arch}.

We train each model for 100 epochs with a batch size of 32 and a learning rate of 0.001 using the Adam optimizer~\cite{ADAM2015}. The parameters at the epoch corresponding to the least validation loss (negative log-likelihood) were selected using training checkpoints and used for evaluation.

\section{Datasets and Preprocessing}
\label{app:data}

\begin{table*}[h!]
\caption{{\bf Dataset details.} List of datasets used for evaluating outlier detection with PixelCNN++. }
\centering
\resizebox{0.5\pdfpageheight}{!}{%
\begin{tabular}{llcrr}
\toprule
{\bf Dataset} & {\bf Type} & {\bf Exemplars} & {\bf N-train (N-val)} & {\bf N-test}\\

\midrule
MNIST & Grayscale & \includegraphics[height=0.35in]{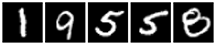} & 54000 (6000) & 10000 \\
Fashion-MNIST & Grayscale & \includegraphics[height=0.35in]{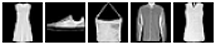} & 54000 (6000) & 10000\\
EMNIST-Letters & Grayscale & \includegraphics[height=0.35in]{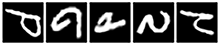} & 79920 (8880) & 14800\\
Sign Language MNIST & Grayscale & \includegraphics[height=0.35in]{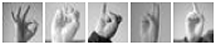} & 24720 (2735) & 7172\\
CLEVR & Grayscale & \includegraphics[height=0.35in]{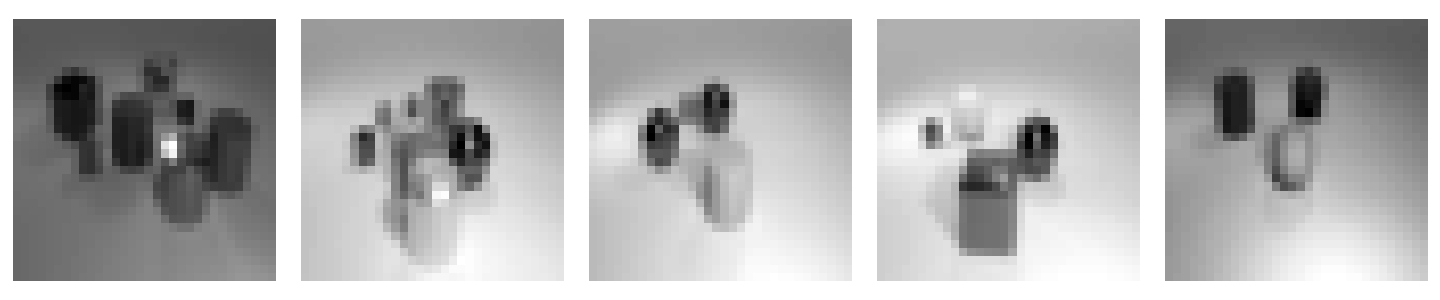} & 63000 (7000) & 15000\\
Gray-Noise & Grayscale & \includegraphics[height=0.35in]{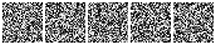} & - & 10000\\
Gray-Constant & Grayscale & \includegraphics[height=0.35in]{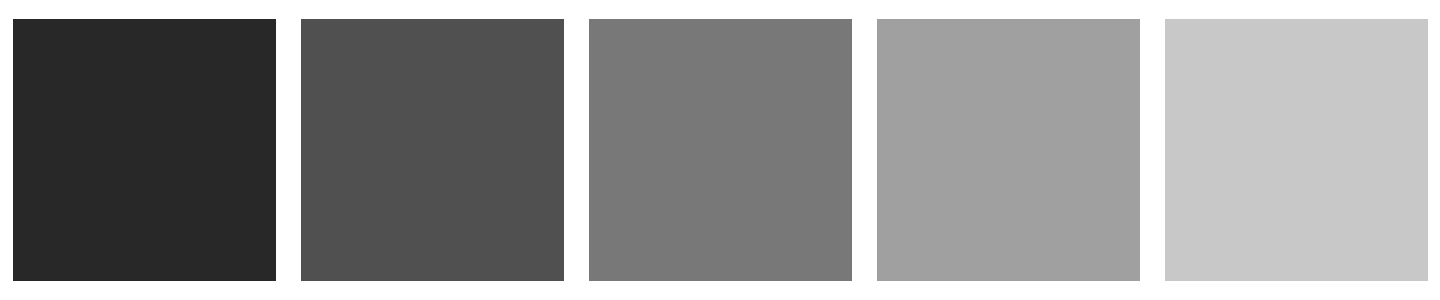} & - & 256\\

\midrule

SVHN & Color & \includegraphics[height=0.35in]{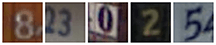} & 65932 (7325) & 26032\\
CelebA & Color & \includegraphics[height=0.35in]{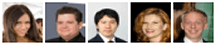} & 146493 (16277) & 19962\\
CompCars & Color & \includegraphics[height=0.35in]{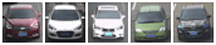} & 28034 (3114) & 13333\\
GTSRB & Color & \includegraphics[height=0.35in]{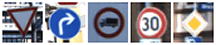} & 35289 (3920) & 12630\\
CIFAR-10 & Color & \includegraphics[height=0.35in]{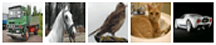} & 45000 (5000) & 10000\\
LSUN-Classroom & Color & \includegraphics[height=0.35in]{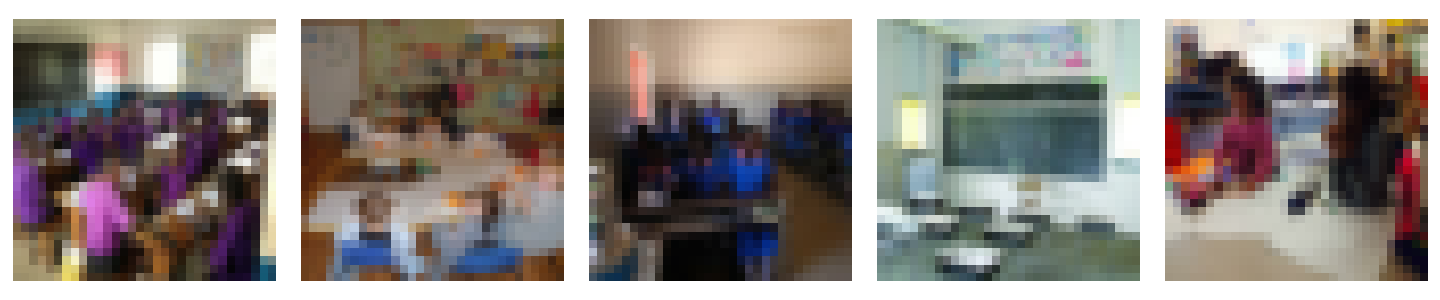} & 134504 (16813) & 16813\\
Color-Noise & Color & \includegraphics[height=0.35in]{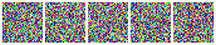} & - & 10000\\
Color-Constant & Color & \includegraphics[height=0.35in]{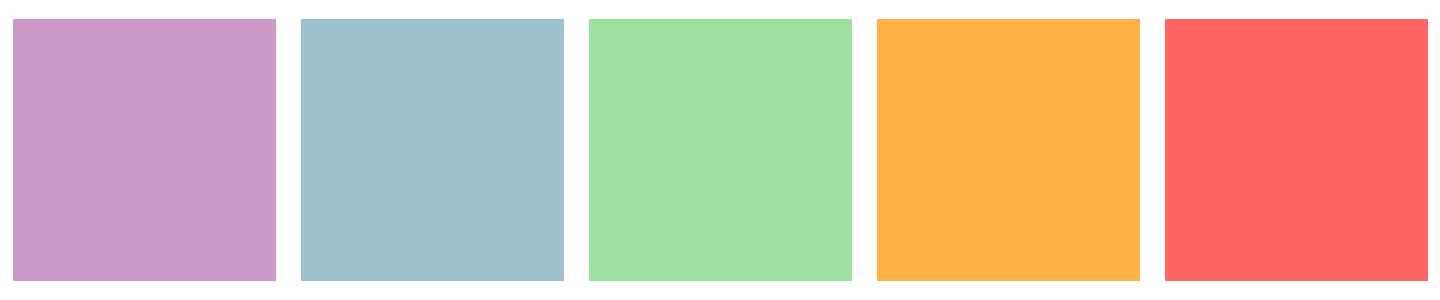} & - & 10000\\
\bottomrule

\end{tabular}
}
\label{tab:data_details}
\end{table*}

Nearly all of our datasets are available in the TensorFlow Datasets\footnote{\url{https://www.tensorflow.org/datasets}} library, except for CelebA~\cite{celeba}, GTSRB~\cite{gtsrb}, CompCars~\cite{compcars}, and Sign Language MNIST\footnote{\url{https://www.kaggle.com/datamunge/sign-language-mnist}}; the latter were manually downloaded from their respective sources. For the EMNIST dataset, we only use the ``letters'' subset, while for LSUN, we use the ``classroom'' (default) subset. All images were loaded as 8-bit images and resized to 32 $\times$ 32 pixels (x $\times$ y). 

We set apart 10\% of the training samples for validation while the evaluation was performed on the designated test splits of each dataset. As the LSUN classroom dataset has only 300 samples in its test set, we set apart another 10\% from the training set (in addition to the validation set) for evaluation. The list of datasets used, along with the number of samples in the train, validation, and test splits, are shown in Table.~\ref{tab:data_details}, along with example images from each dataset.

\section{De-biasing PixelCNN++ likelihoods: Additional results}

\subsection{Bias in PixelCNN++ likelihoods for grayscale images}
\label{app:lik_components}

\begin{figure}[thb]
    \centering
    \includegraphics[width=0.8\linewidth]{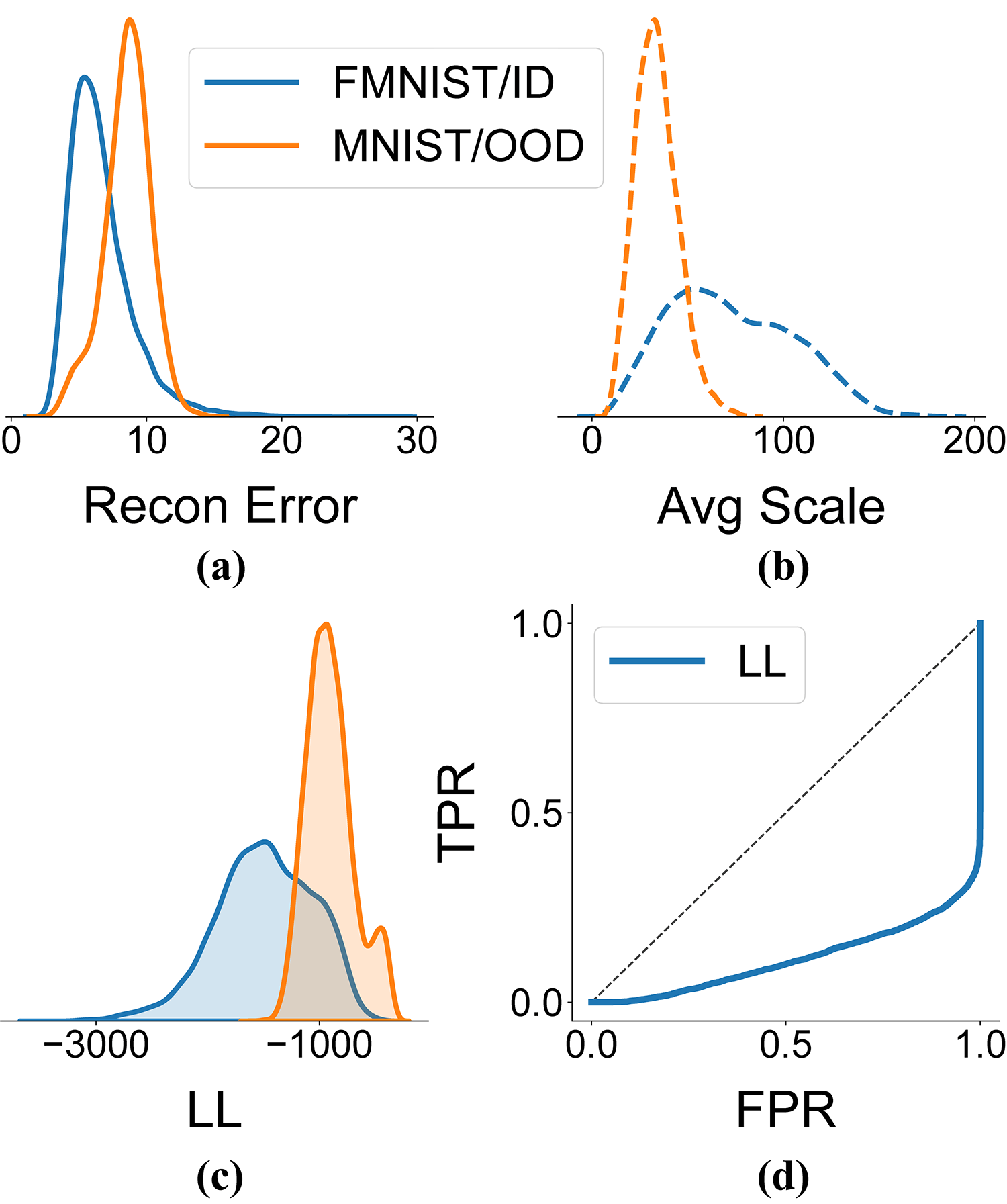}
    \caption{\textbf{Biases in PixelCNN++ likelihoods for grayscale images. (a-d)} Same as in Figures.~\ref{fig1}e-h, respectively, but for FMNIST/ID (blue) vs MNIST/OOD (orange).}
    \label{app:fig:ll_components}
    \vspace{-1em}
\end{figure}

In section~\ref{sec:lik_components}, we analyzed the origins of bias in PixelCNN++ likelihoods for natural image datasets  (CIFAR10/ID vs SVHN/OOD). We observed that the reconstruction error and scale were lower for the OOD samples, and the log-likelihood was higher for OOD samples than for ID samples, leading to anomalous outlier detection AUROCs (Fig.~\ref{fig1}h). Here, we analyze another well-reported  challenging case for outlier detection with grayscale images -- FashionMNIST (or FMNIST)/ID vs MNIST/OOD. 

In this case, we found that reconstruction errors were marginally lower for FMNIST/ID images (Fig.~\ref{app:fig:ll_components}a blue density) as compared to the MNIST/OOD images (Fig.~\ref{app:fig:ll_components}a, orange density). Yet, the scale was markedly lower for MNIST/OOD images (Fig.~\ref{app:fig:ll_components}b, orange density) than for the FMNIST/ID images (Fig.~\ref{app:fig:ll_components}b, blue density). Despite higher reconstruction errors, the lower scale sufficed to yield higher likelihoods for MNIST/OOD samples (Fig.~\ref{app:fig:ll_components}c, orange density) as compared to FMNIST/ID samples (Fig.~\ref{app:fig:ll_components}c, blue density); this resulted in anomalous outlier detection AUROC curves (Fig.~\ref{app:fig:ll_components}d, bowed downwards). 

In other words, more confident PixelCNN++ predictions for MNIST/OOD as compared to FMNIST/ID resulted in anomalous likelihoods for OOD samples' poor outlier detection. The greater confidence is likely due to the large number of black background pixels in the MNIST dataset  ~\cite{Ren2019}. The PixelCNN++ model can readily predict these background pixels simply by examining a local neighborhood of preceding pixel values. 

\subsection{Comparing PixelCNN++ likelihoods and input complexity (IC): Methods}
\label{app:ic}

\begin{figure}[thb]
\centering
\includegraphics[width=\linewidth]{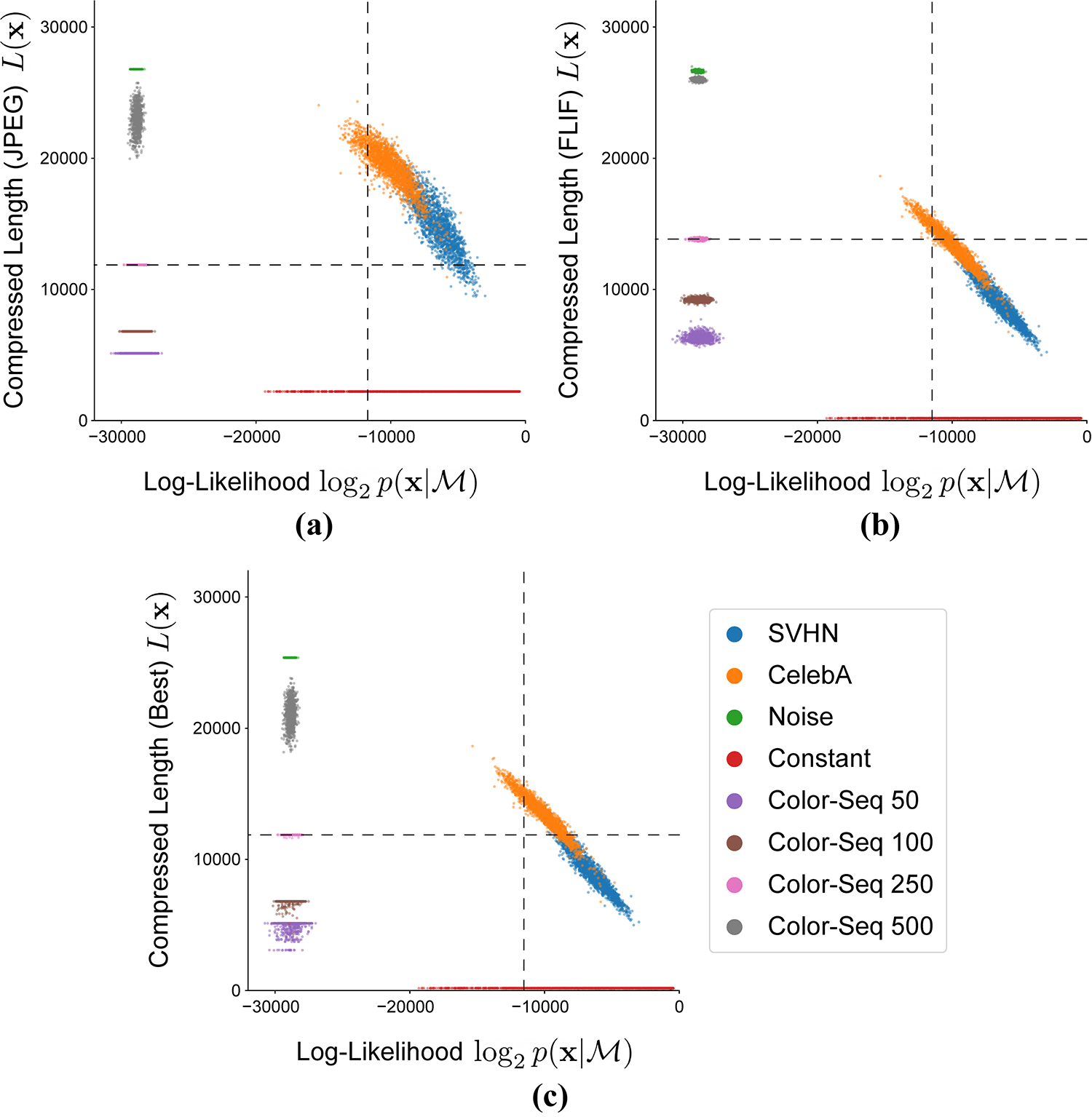}
\caption{\textbf{Compressed lengths have a many-to-one relationship with PixelCNN++ likelihoods across compression techniques.} Same as in Figure~\ref{fig2} but by using \textbf{(a)} JPEG, \textbf{(b)} FLIF, and \textbf{(c)} Best compressors for computing the compressed lengths.}
\label{app:fig:ic_fig}
\vspace{-1em}
\end{figure}

In section~\ref{sec:ic} (main text), we used constant, noise, and repeating color sequence images to show the mapping between PixelCNN++ likelihoods and compressed lengths. These images were all 3 channel images and were generated as follows:

\textbf{Constant:} These represent uniformly colored images. Three random integers were independently sampled from a uniform distribution in the range [0,255] -- one for each of the red, green, and blue channels. Each channel assumes the same value across all pixels in the image, thereby producing a uniform-colored 8-bit image.

\textbf{Noise:} These represent noise images with independent sub-pixel. Each sub-pixel was randomly and independently sampled from a uniform distribution in the range [0,255].

\textbf{Color Sequence:} These represent images with periodic sequences of colors. For example, a Color-Seq 50 image contained 50 randomly sampled colors arranged as a repeating sequence, and a Color-Seq 2 is an alternating checkerboard pattern. More generally, for color-seq K, K random independent integers were sampled from a uniform distribution in the range [0,255] for each channel (3K integers in all). For each channel, its respective K intensities were placed in a raster scan order (row first), followed by repeating the same sequence of values until the end of the image was reached. This produced a repeating sequence of K colors in the image. 

Using these images, we showed that the mapping between PixelCNN++ likelihoods and PNG compressed lengths was not even approximately linear. Here, we show that our claims are consistent with other compressors like JPEG (Fig.~\ref{app:fig:ic_fig}a) and FLIF  (Fig.~\ref{app:fig:ic_fig}b). We also observed similar results when considering the ``best'' compressor (Fig.~\ref{app:fig:ic_fig}c), which is computed as the minimum size for a given image across the three compressors $L(\mathbf{x})$= min($L_1(\mathbf{x})$, $L_2(\mathbf{x})$, $L_3(\mathbf{x})$)). 

In summary, regardless of the compressor used, widely different log-likelihoods occurred for images with identical compressed lengths (Fig.~\ref{app:fig:ic_fig}, dashed horizontal lines) and, conversely, images with closely similar log-likelihoods exhibited widely different compressed lengths (Fig.~\ref{app:fig:ic_fig}, dashed vertical lines).

\subsection{PixelCNN++ ablation experiments: Methods}
\label{app:local_dep}

\begin{figure}[!ht]
\centering
\includegraphics[width=\linewidth]{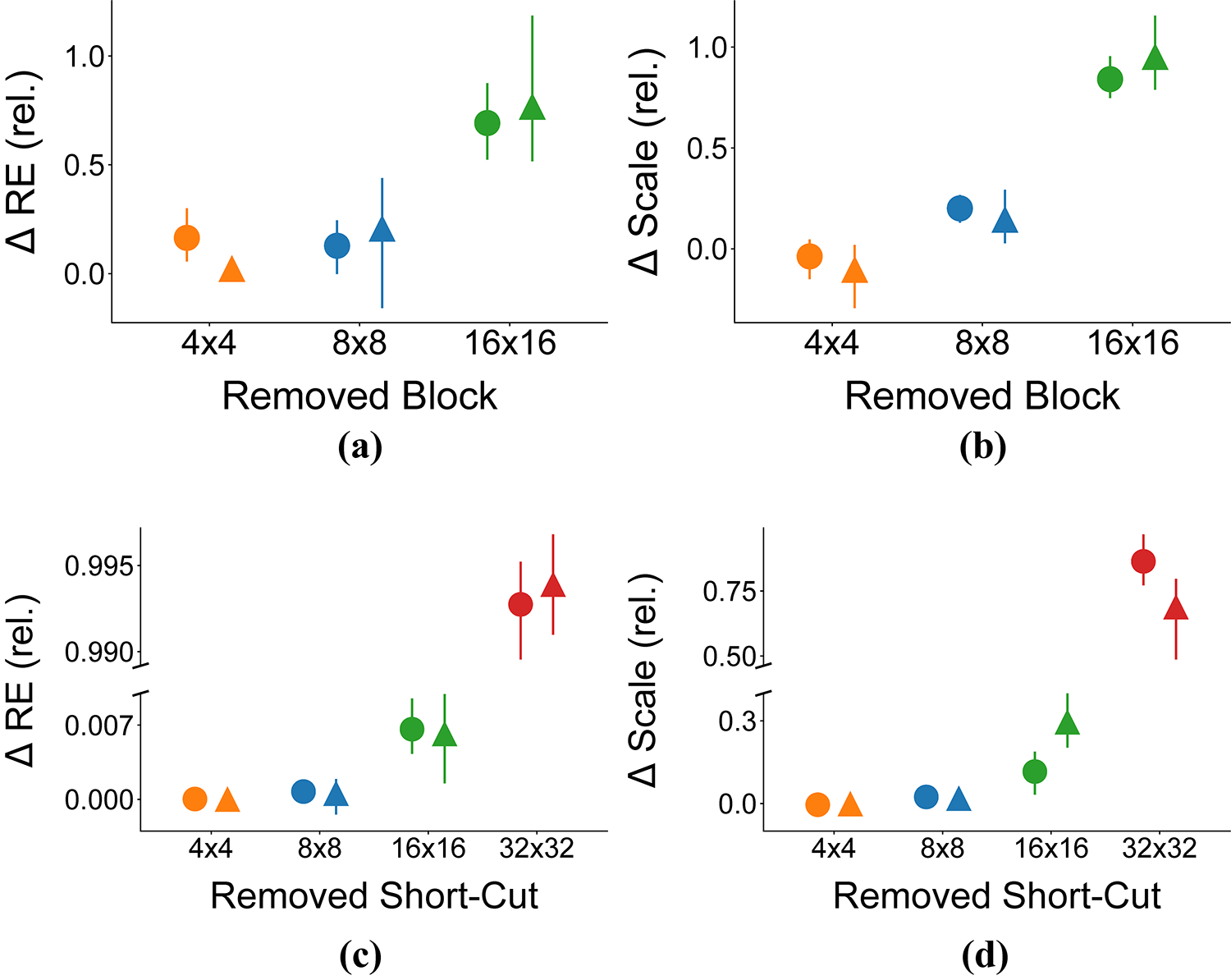}
\caption{\textbf{PixelCNN++ relies heavily on local dependencies for prediction.} \textbf{(a, b)} Same as in Figure~\ref{fig3}b but showing the effect of block ablations on reconstruction error and scale respectively. \textbf{(c, d)} Same as in Figure~\ref{fig3}c but showing the effect of short-cut ablations on reconstruction error and scale, respectively.}
\label{app:fig:ablation_exp}
\vspace{-1em}
\end{figure}

We performed two kinds of ablation experiments: a) by removing the symmetric convolutional blocks (or hierarchies) of the PixelCNN++ network (Fig.~\ref{fig3}a, boxes), and b) by removing the short-cut connections (Fig.~\ref{fig3}a, curved, dashed arrows). We quantified the effect of each type of ablation on model likelihoods. 

For the first series of experiments (a), we sought to measure the contribution of the 4$\times$4, 8$\times$8, and 16$\times$16 blocks to the pixel level predictions in terms of both the location and scale parameters. Although we could not remove the 8$\times$8 or 16$\times$16 layers individually due to the nested (sequential) layout of these layers, we quantified their contribution to the model log-likelihood (LL) as follows: the contribution of the $K \times K$ layers was quantified as the change in LL upon removing all innermost layers up to size $K \times K$ minus the change in LL upon removing all innermost layers up to size $K/2 \times K/2$. For example, the contribution of the 8$\times$8 layers was quantified as the change in LL upon removing the two (4$\times$4 and 8$\times$8) innermost layers minus the change in LL upon removing the 4$\times$4 layer alone. The outermost (32$\times$32) block could not be removed because that would entail ablating virtually the entire PixelCNN++ network (Fig.~\ref{fig3}a). Each of these changes in LL was normalized (divided) by the changes in LL upon removing all 3 sets of blocks to quantify the relative magnitude of the contribution of each block; the normalization was done separately for the ID and OOD data.

For the second series of experiments (b), we ablated short-cut connections across each symmetric pair of blocks (4$\times$4, 8$\times$8, 16$\times$16, and 32$\times$32) independently of the other connections (Fig.~\ref{fig3}a, curved, dashed arrows). In this case, also, changes in LL were normalized (divided) by the changes in LL upon removing all 4 sets of short-cut connections to quantify the relative magnitude of the contribution of each set of connections. Note that in Figure~\ref{fig3}c, there is a break in the y-axis indicating the inordinately high effect on LL upon removing the 32$\times$32 block's short-cut connections.

In Figures~\ref{fig3}b-c, we showed the effects of ablation on log-likelihoods. Here, we also show that these changes in likelihoods are accompanied by progressively higher reconstruction errors (Fig.~\ref{app:fig:ablation_exp}a, c) and more uncertain predictions (Fig.~\ref{app:fig:ablation_exp}b, d). As with the likelihoods, the largest effects on reconstruction error and scale occur for the outermost layers (16$\times$16 for block ablation and 32$\times$32 for short-cut connection ablation), which capture the most local dependencies in the model.

\subsection{``Stirring'' and ``Shaking'': Motivation}
\label{app:stirring}

\begin{figure}[thb]
\centering
\includegraphics[width=\linewidth]{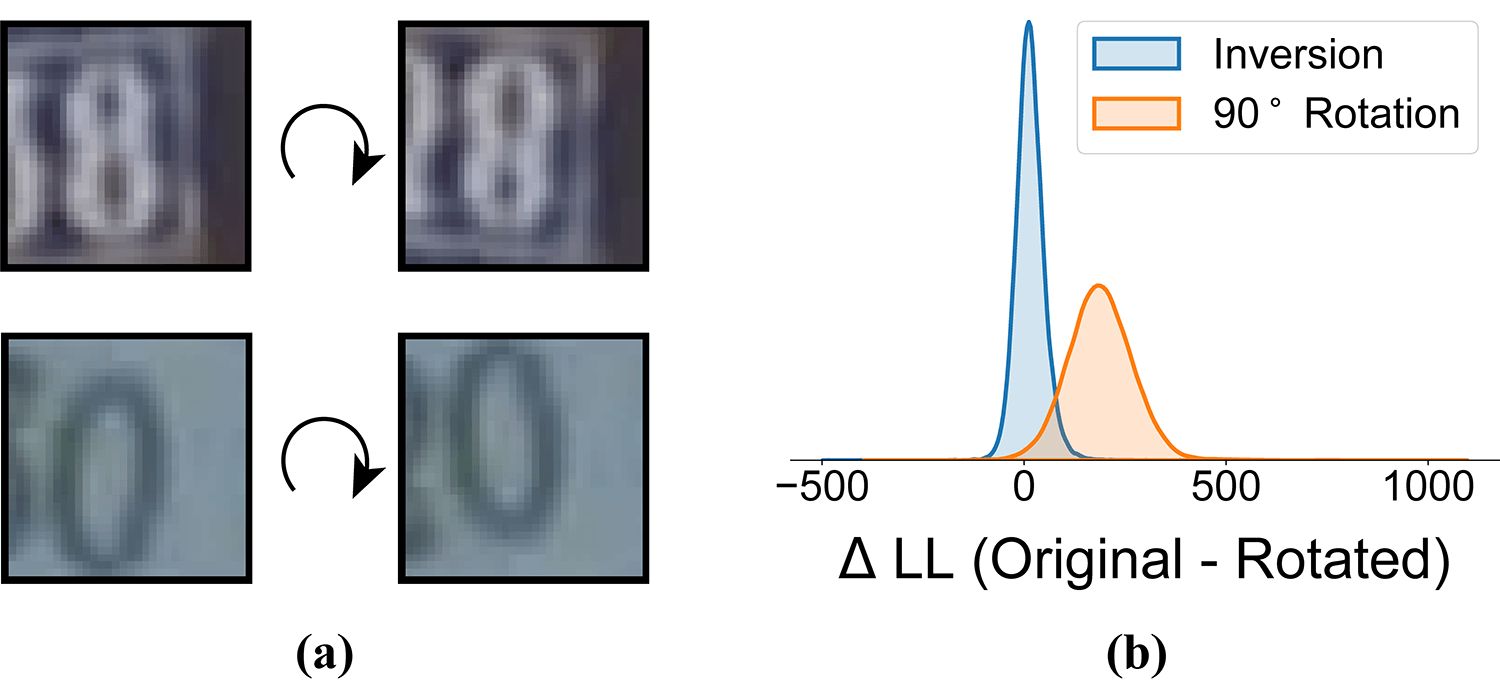}
\caption{\textbf{Motivation for ``Stirring''.} \textbf{(a)} Exemplars from the SVHN dataset before and after inversion illustrate a horizontal axis of near-symmetry in these images. \textbf{(b)} Difference between the log-likelihoods of the original and transformed SVHN/ID images with inversion (blue) and with 90$^\circ$ rotation (orange).}
\label{app:fig:stirring}
\vspace{-1em}
\end{figure}

\begin{figure}[thb]
\centering
\includegraphics[width=0.55\linewidth]{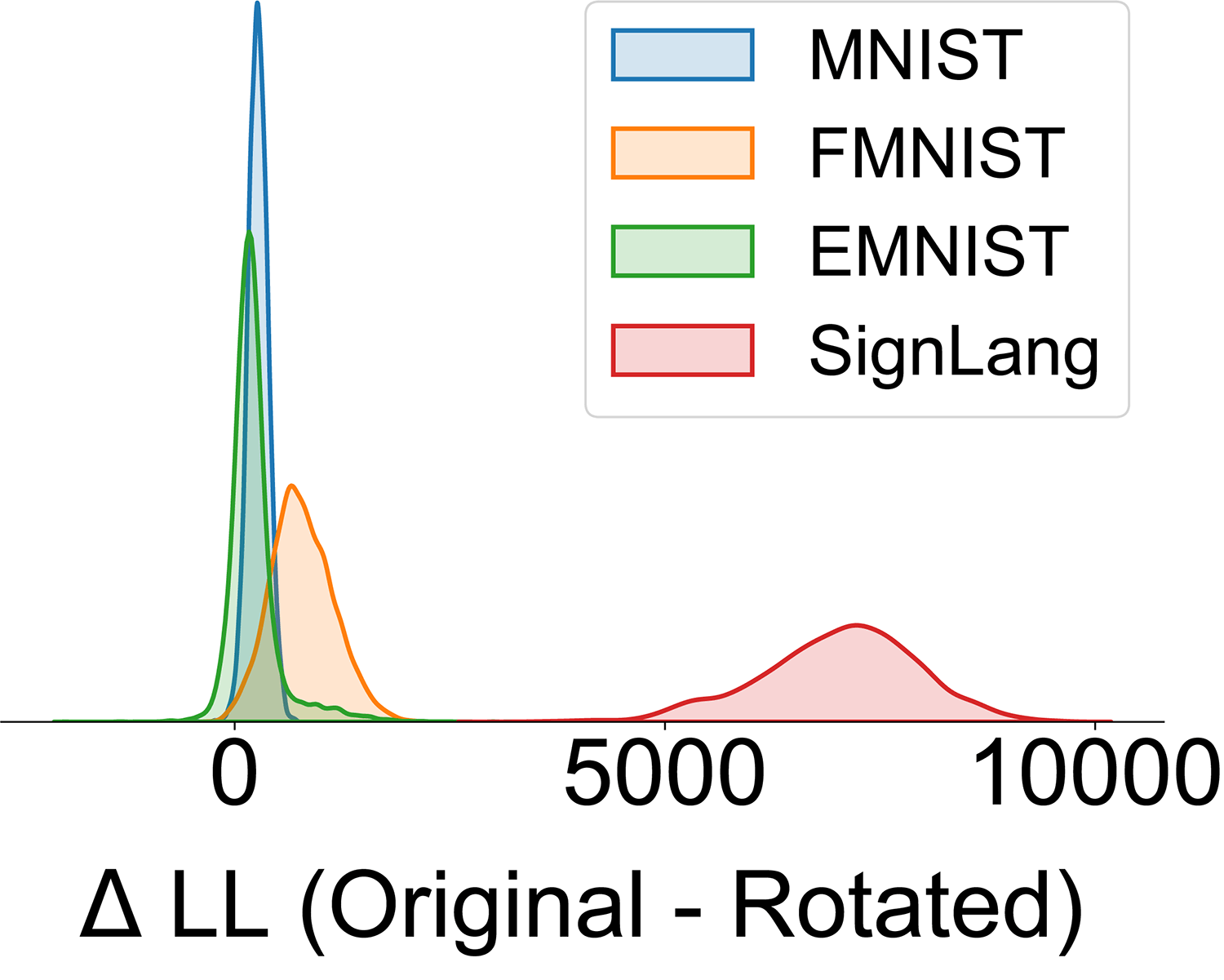}
\caption{\textbf{Motivation for ``Shaking''.} $\log p_{\mathrm{LR}}$ based on ``stirring'' for MNIST/ID (blue), EMNIST/ID (green), FashionMNIST/ID (orange) and SignLang/ID (red), with PixelCNN++ models trained on the respective datasets.}
\label{app:fig:shaking}
\vspace{-1em}
\end{figure}

{\bf Motivation for ``stirring''.} In the main text, we showed how likelihoods for inverted faces were substantially lower than those for upright faces for a PixelCNN++ model trained on the CelebA dataset (Fig.~\ref{fig3}d-e). We also showed how this was likely a result of the model having learned long-range dependencies in the ID, but not OOD, data.

While inversion works well to isolate long-range dependencies for images that have a natural, upright orientation (e.g., faces or cars), such a perturbation may not work well for all datasets. For example, in datasets like SVHN, images of certain digits may have a clear, horizontal axis of symmetry (e.g., 0-s, 1-s or 8-s) (Fig.~\ref{app:fig:stirring}a). Or certain digits may resemble others when inverted (e.g., 6-s and 9-s). In these cases, if the inverted inlier image is substantially visually similar to an upright inlier image, then the PixelCNN++ model would yield similar likelihoods for both inverted and upright inlier images. Indeed this is what we observed when we compared PixelCNN++ likelihoods for inverted and upright images with a model trained on the SVHN dataset (Fig.~\ref{app:fig:stirring}b, blue density); note that mode of the $\Delta$ LL(=LL-upright - L-inverted) density is around zero. To ameliorate this challenge, one may then consider a different rotational transformation that is not vulnerable to this symmetry. For example, a 90$^{\circ}$ clockwise rotation yielded significantly lower likelihoods for the rotated inlier images than for upright images in the SVHN dataset (Fig.~\ref{app:fig:stirring}b, orange density); note that mode of the $\Delta$LL density is clearly greater than zero. Consequently, we considered a full family of transformations, including rotation by different angles of the upright and the flipped images. This collection of transformation is what we call ``stirring''.

{\bf Motivation for ``shaking''.} Despite ``stirring'', some types of images may comprise features of such simplicity that a powerful PixelCNN++ model trained on upright images alone could reconstruct stirred images accurately and confidently. We note that this happens primarily in simplistic grayscale datasets like MNIST and EMNIST but not for more complex grayscale image datasets like FashionMNIST and SignLanguageMNIST (Fig.~\ref{app:fig:shaking}). For these datasets, likelihoods for the original and stirred images largely overlap, likely because edges, contours, and other simple features associated with digits and letters are simple enough for the PixelCNN++ model to predict, regardless of the orientation of the images. In these cases, estimating the contribution of long-range dependencies to the log-likelihood ($\log p_{\mathrm{LR}}$) becomes challenging with ``stirring''. Consequently, we consider a second class of bijective transformations that involve dividing the images into patches and randomly shuffling these patches spatially. 

Specifically, we consider the following schemes for patching and shuffling the images: i) 1 derangement by splitting the image in half along the horizontal mid-line, ii) 1 derangement by splitting the image in half along the vertical mid-line, iii) 9 derangements by splitting the image into four quarters and shuffling. 2 derangements in scheme (iii) are redundant with the derangements in (i) and (ii). Thus, we obtain a total of 9 unique derangements across the 3 schemes. This collection of derangements is what we call ``shaking''.

\subsection{Conditional Correction: Methods}
\label{app:cond_corr}

\begin{figure}[thb]
\centering
\includegraphics[width=0.7\linewidth]{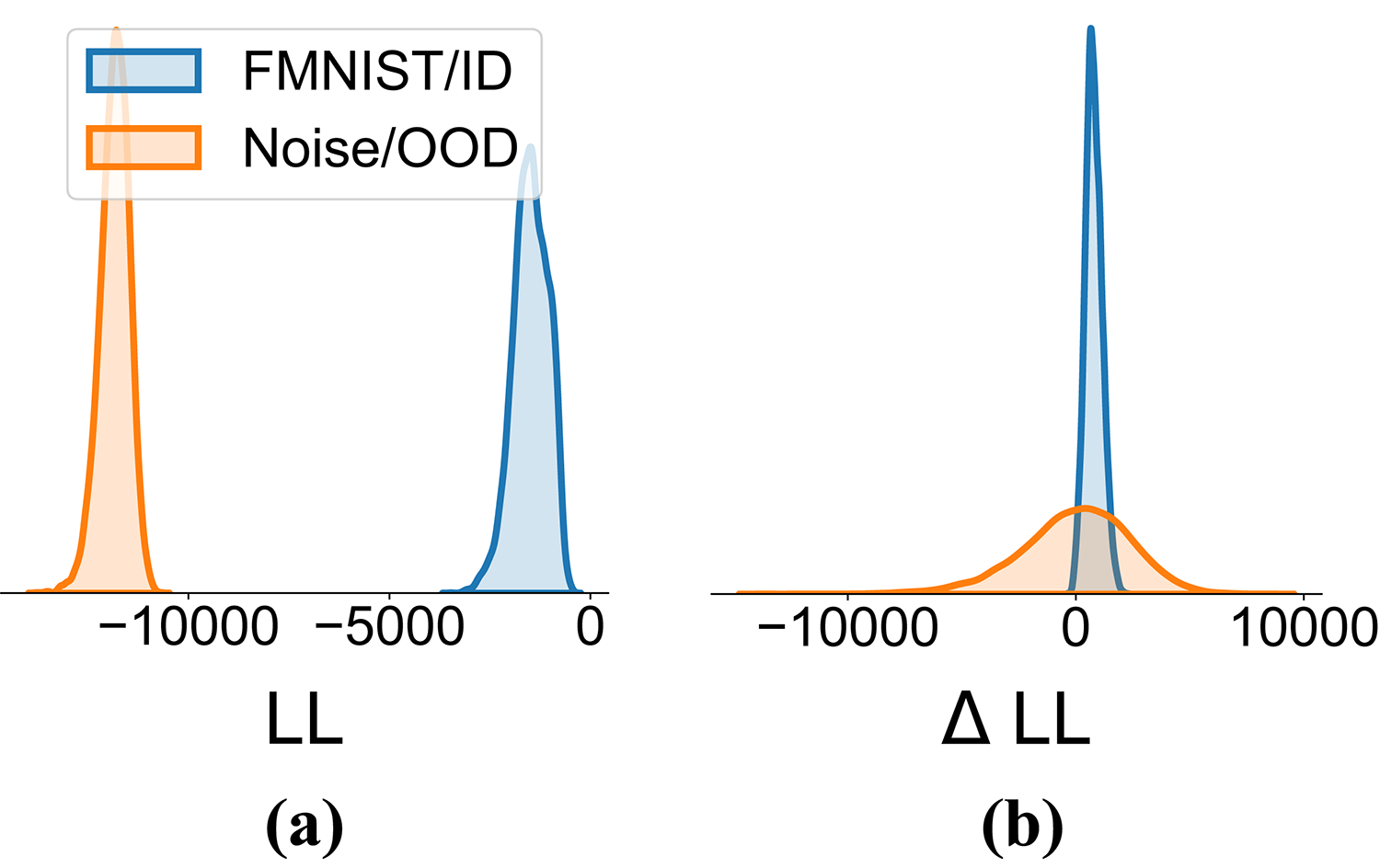}
\caption{\textbf{Motivation for conditional correction.} \textbf{(a)} Log likelihood of Noise/OOD samples (orange) and FMNIST/ID samples (blue) for a PixelCNN++ trained on the FMNIST dataset. \textbf{(b)} $\log p_{\mathrm{LR}}$ based on ``stirring'' for Noise/OOD samples (orange) and for FMNIST/ID samples (blue).}
\label{app:fig:cond_corr}
\vspace{-1em}
\end{figure}

The $\log p_{\mathrm{LR}}$ based on either ``stirring'' or ``shaking'' can be a noisy estimate, especially when the likelihoods involved are very small numbers. We illustrate this with an example of a PixelCNN++ model trained on the FMNIST dataset; we observe that the log-likelihood for Noise/OOD is many orders of magnitude smaller than that for FMNIST/ID (Fig.~\ref{app:fig:cond_corr}a). As a result, the $\log p_{\mathrm{LR}}$ based on ``stirring'' has a much larger variance across samples for the Noise/OOD than for the FMNIST/ID dataset (Fig.~\ref{app:fig:cond_corr}b).

To ameliorate this challenge, we adopt the following rational strategy for outlier detection. We first identify outliers below the left-tail of the log-likelihood distribution of the training data as being below 3 times the median absolute deviation (3-MAD) of the median~\cite{leys2013}. When a test sample's log-likelihood falls below the 3-MAD criterion, it is directly labeled as an outlier. After filtering based on this condition, only samples whose log-likelihood overlaps with the training log-likelihood distribution are considered for our corrections with the ``stirring'' and ``shaking'' transformations. 

All of the outlier detection results with ``stirring'' and ``shaking'' reported in the main text (Figs.~\ref{fig:grayscale} and~\ref{fig:color}) include the conditional correction; results without the conditional correction are shown in Appendix~\ref{app:without_cond}.

\section{Outlier Detection: Additional Results}
\label{app:other_metrics}

\begin{figure*}[!ht]
    \centering
    \begin{subfigure}[b]{0.95\textwidth}
        \centering
        \includegraphics[width=0.95\linewidth]{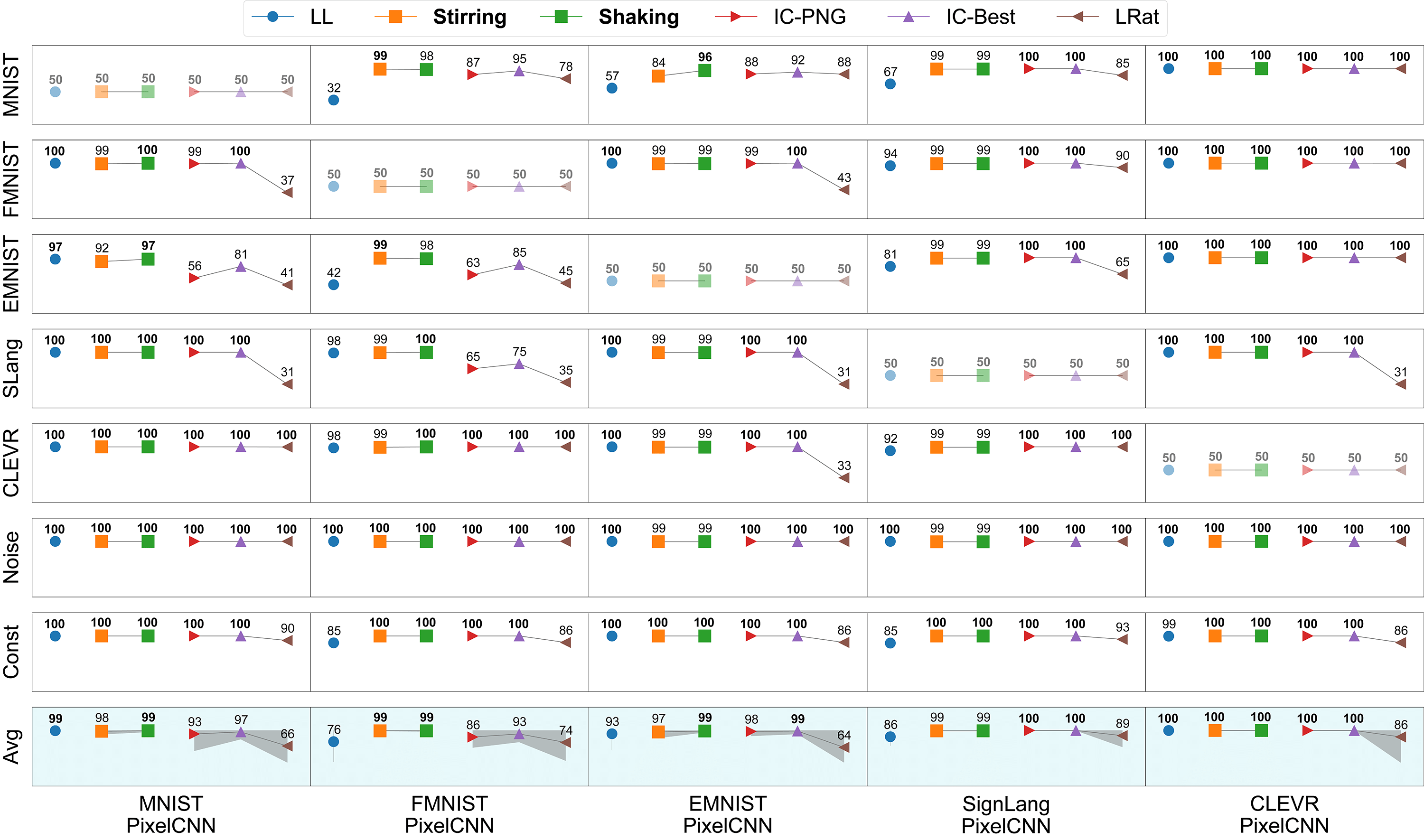}
        \caption{}
        \label{app:fig:grayscale_auprc}
    \end{subfigure}
    \begin{subfigure}[b]{0.95\textwidth}
        \centering
        \includegraphics[width=0.95\linewidth]{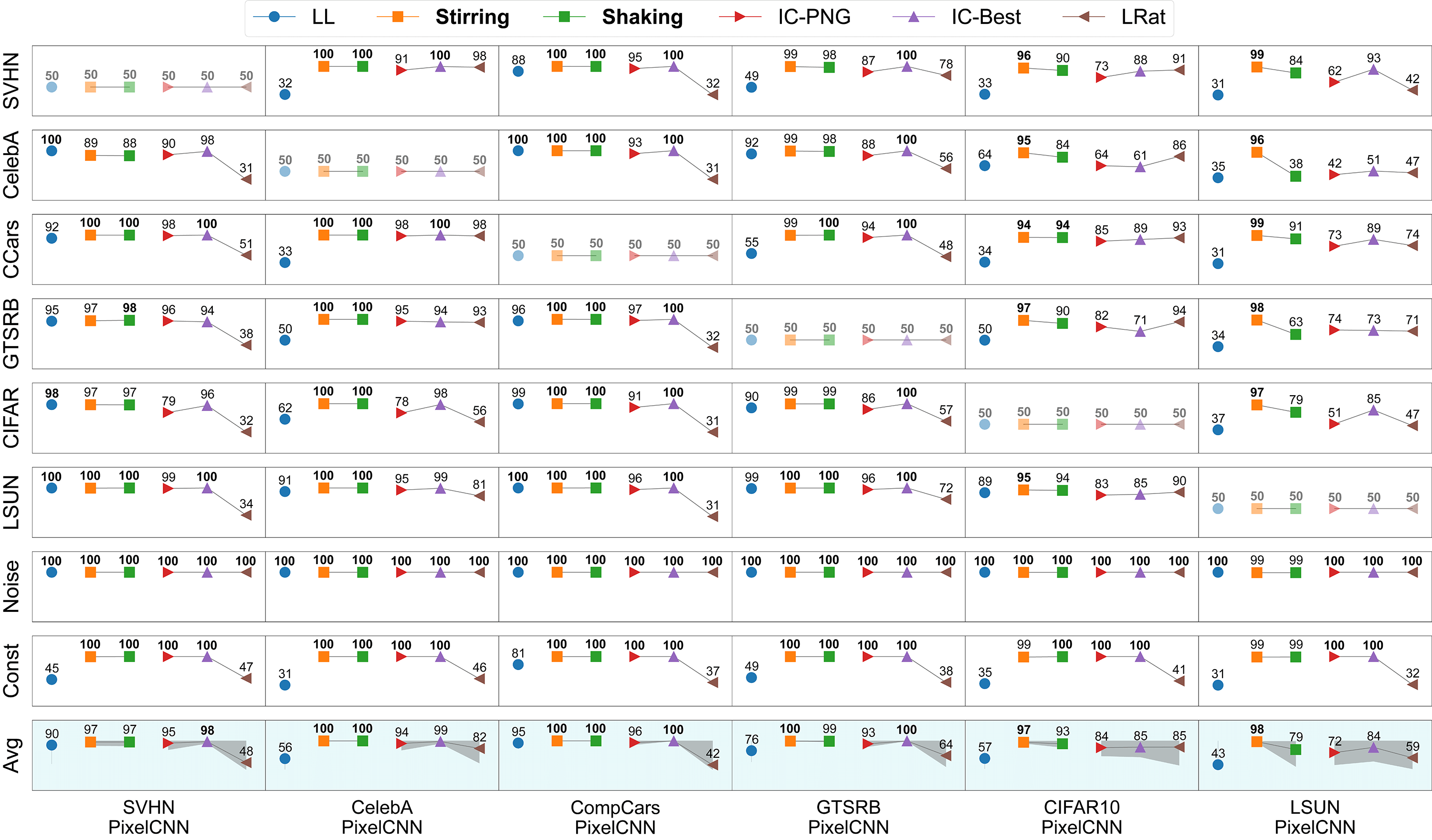}
        \caption{}
        \label{app:fig:color_auprc}
    \end{subfigure}
    \caption{\textbf{Outlier detection performance measured using the area under the precision-recall curve.} \textbf{(a)} Same as in Figure~\ref{fig:grayscale}, but results are shown using the area under the precision-recall curve (AUPRC) instead of AUROC. \textbf{(b)} Same as in Figure~\ref{fig:color} but results are shown using AUPRC instead of AUROC. Higher values indicate better outlier detection performance. Other conventions are the same as in Figures~\ref{fig:grayscale}-\ref{fig:color}.}
    \vspace{-1em}
\end{figure*}

\begin{figure*}[!ht]
    \centering
    \begin{subfigure}[b]{0.95\textwidth}
        \centering
        \includegraphics[width=0.95\linewidth]{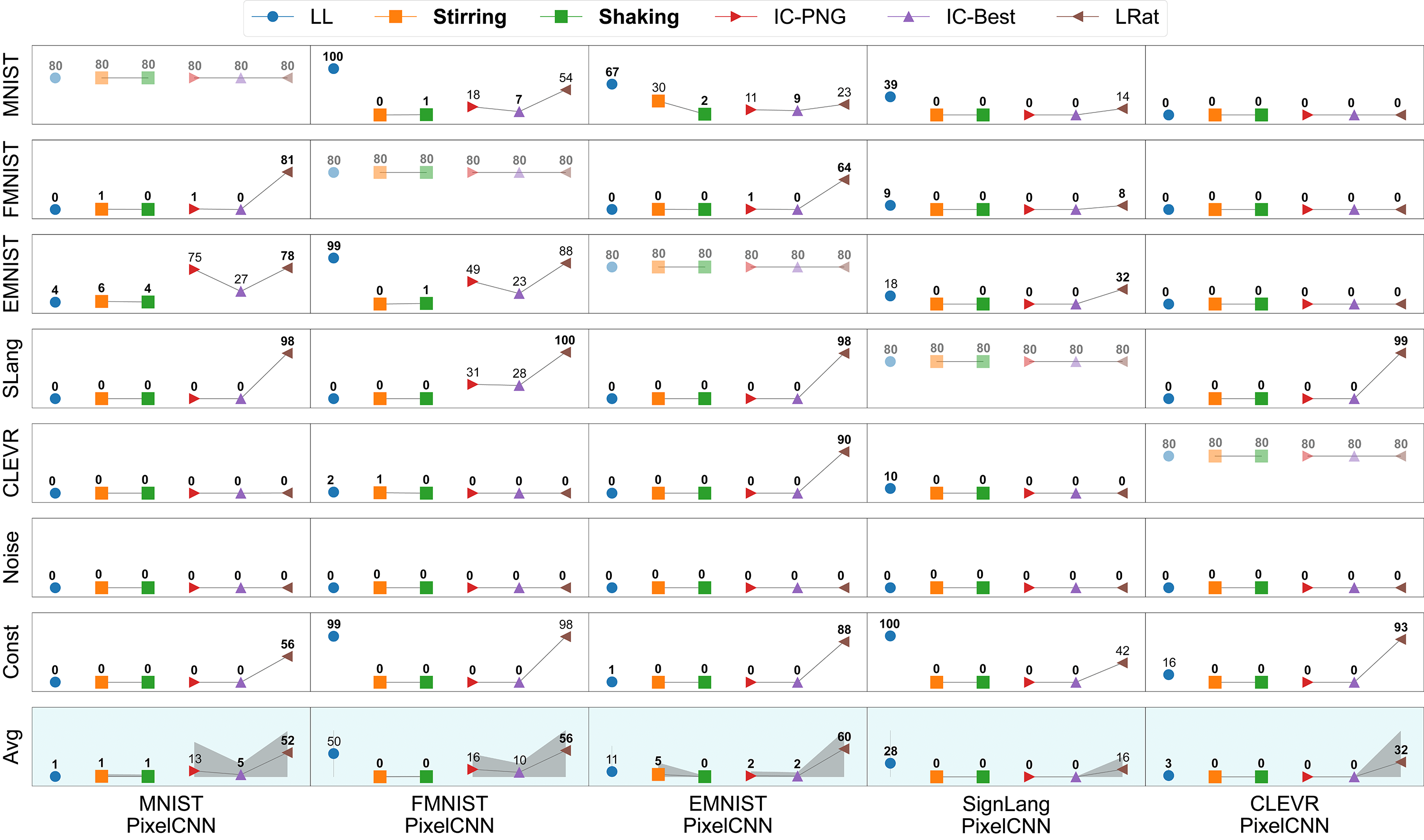}
        \caption{}
        \label{app:fig:grayscale_fpr80}
    \end{subfigure}
    \begin{subfigure}[b]{0.95\textwidth}
        \centering
        \includegraphics[width=0.95\linewidth]{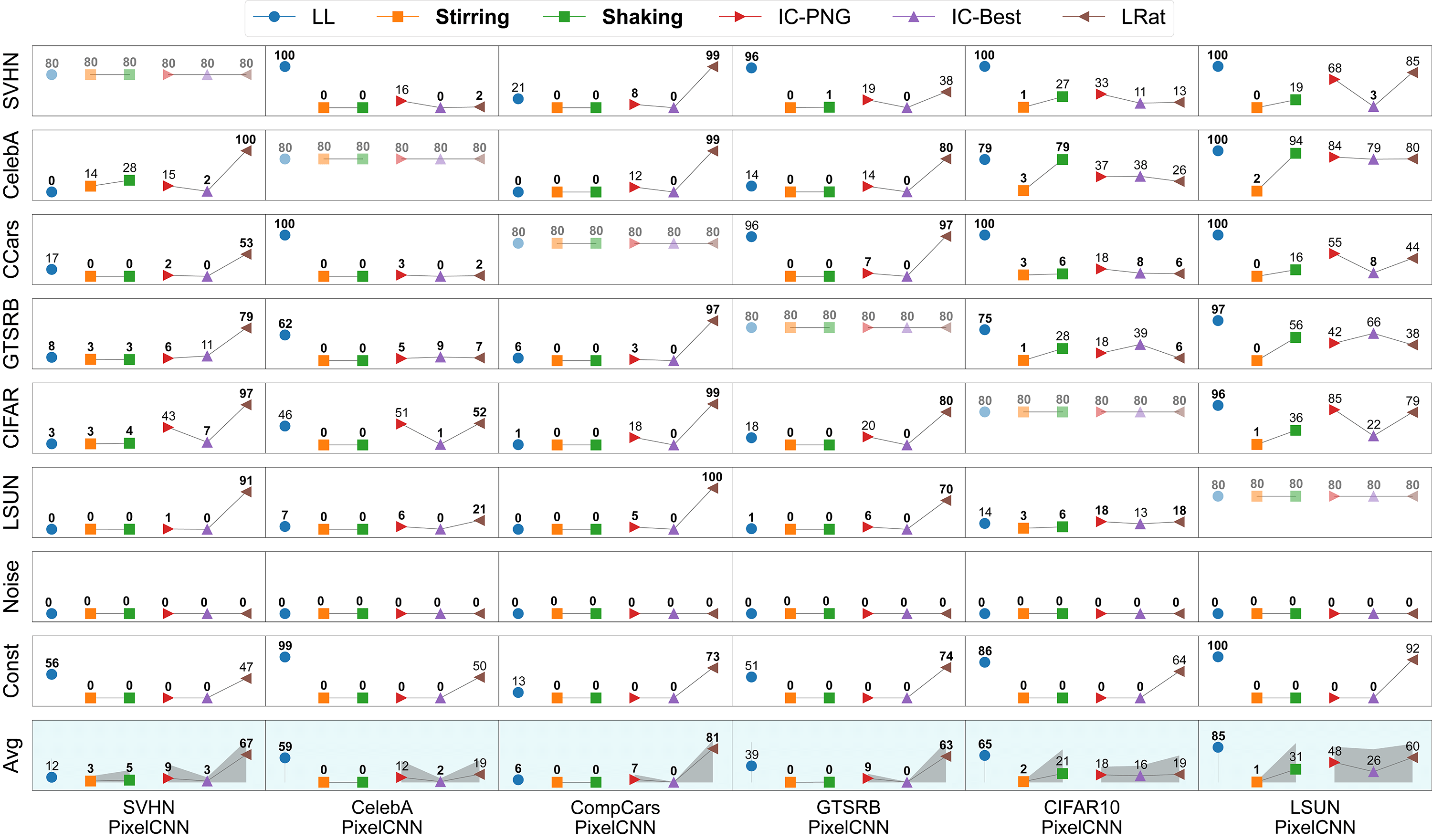}
        \caption{}
        \label{app:fig:color_fpr80}
    \end{subfigure}
    \caption{\textbf{Outlier detection performance measured using false-positive rate at 80\% true-positive rate.} \textbf{(a)} Same as in Figure~\ref{fig:grayscale} but results are shown using false-positive rate at 80\% true-positive rate (FPR@80\%TPR) instead of AUROC. \textbf{(b)} Same as in Figure~\ref{fig:color} but results are shown using FPR@80\%TPR instead of AUROC. Lower values indicate better outlier detection performance. Other conventions are the same as in Figures~\ref{fig:grayscale}-\ref{fig:color}.}
    \vspace{-1em}
\end{figure*}

The results in the main text quantify outlier detection performance using Area Under the Receiver Operating Characteristic (AUROC) curve (Figs.~\ref{fig:grayscale} and~\ref{fig:color}). The ROC is a plot of true-positive rate versus false-positive rates, and a higher AUROC indicates better outlier detection. All AUROC results are shown for distinguishing ID test versus OOD test samples. For uniformity across all comparisons, we sampled 5000 images randomly from the test split of each of the datasets and computed the AUROC; random seeds and code are available for reproducing these results (see Appendix ~\ref{app:rep_check}).

Here, we show outlier detection performance using the area under the precision-recall curve (AUPRC) (Fig.~\ref{app:fig:grayscale_auprc} for grayscale images and Fig.~\ref{app:fig:color_auprc} for natural images) and false-positive rate at 80\% true-positive rate (FPR@80\%TPR) (Fig.~\ref{app:fig:grayscale_fpr80} for grayscale images and Fig.~\ref{app:fig:color_fpr80} for natural images). Higher values of AUPRC, and lower values of FPR@80\%TPR, indicate better outlier detection. All metrics were calculated using the scikit-learn\footnote{\url{https://scikit-learn.org/stable/}} library.

Similar to the results based on AUROC, these metrics (AUPRC and FPR@80\%TPR) also show that ``stirring'' or ``shaking'' shows substantial improvements over vanilla log-likelihoods for outlier detection.

\section{Comparison with Competing Approaches}
\label{app:eval}

For evaluating competing approaches (Input Complexity and Likelihood Ratios) we use the same PixelCNN++ architecture as mentioned in Appendix ~\ref{app:architecture}.

\begin{figure*}[hbt]
    \centering
    \begin{subfigure}[b]{0.49\textwidth}
        \centering
        \includegraphics[width=0.95\linewidth]{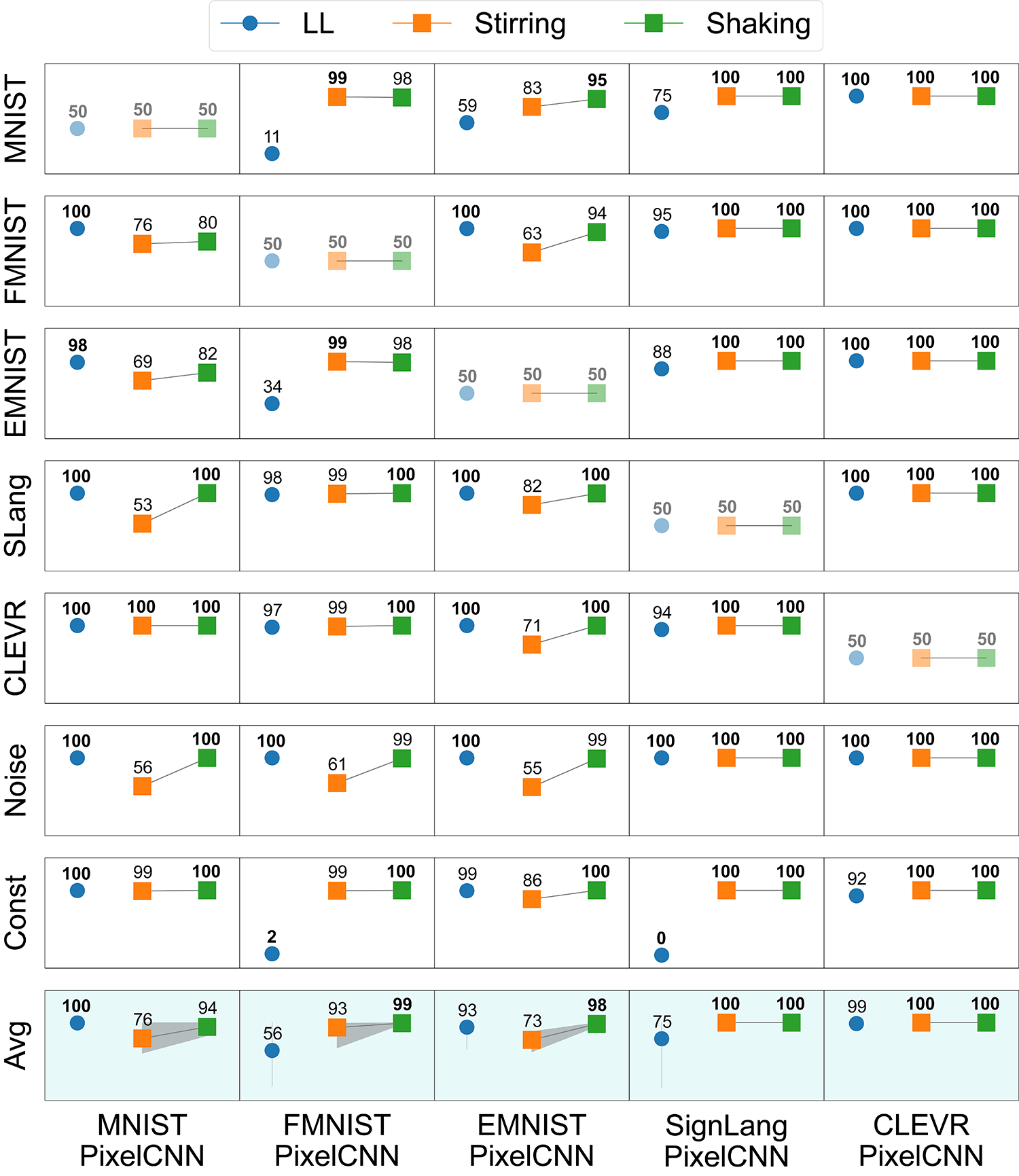}
        \caption{}
        \label{app:fig:grayscale_without_cond}
    \end{subfigure}
    \begin{subfigure}[b]{0.49\textwidth}
        \centering
        \includegraphics[width=0.95\linewidth]{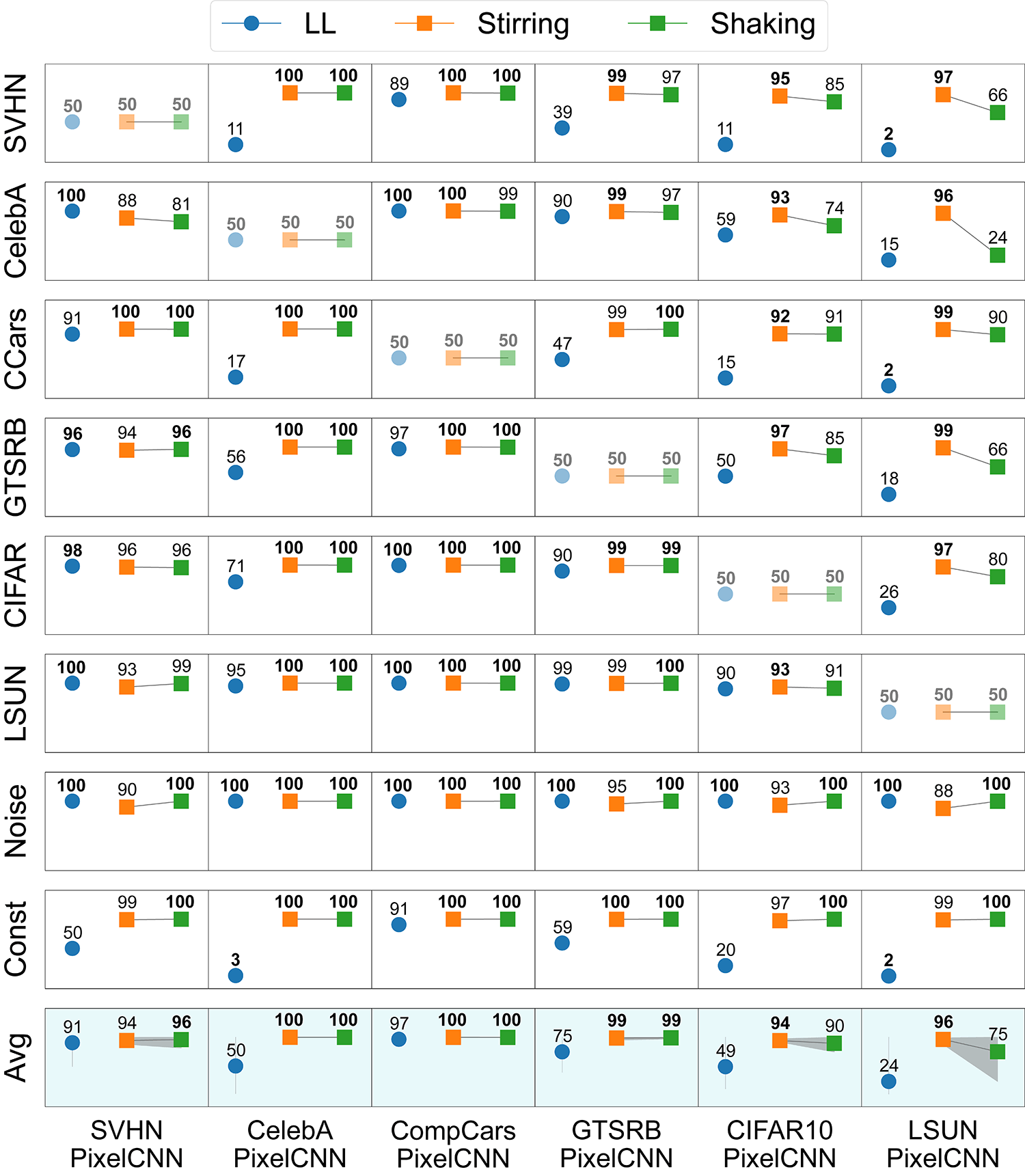}
        \caption{}
        \label{app:fig:color_without_cond}
    \end{subfigure}
    \caption{\textbf{Outlier detection AUROC without conditional correction.} \textbf{(a)} Grayscale image datasets. Blue: log likelihood (LL), uncorrected; Orange: Stirred LL without conditional correction; Green: Shaken LL without conditional correction.  \textbf{(b)} Same as in panel (a) but for natural image datasets. Other conventions are the same as in Figures~\ref{fig:grayscale}-\ref{fig:color}.}
    \label{app:fig:without_cond}
    \vspace{-1em}
\end{figure*}

\textbf{Likelihood Ratio:} The likelihood ratio \cite{Ren2019} was computed using our standard PixelCNN++ models as the foreground models. The background models were trained by corrupting the input data with uniform random noise with a mutation rate $\mu$=0.1 for grayscale datasets and $\mu=0.3$ for natural image datasets, following the recommendation of the original study. The OOD detection score was then computed as the difference between the log likelihood assigned by the foreground model and that assigned by the background model for the original test sample (Figs.~\ref{fig:grayscale} and~\ref{fig:color}, brown symbols). 

Our implementation of likelihood ratio yielded slightly worse AUROC values than those reported by \cite{Ren2019} (Table~\ref{tab:lrat}, first and second rows), possibly because their study employed a more complex PixelCNN++ architecture. AUROC values with ``stirring'' (our method) are, nevertheless, comparable to state-of-the-art values reported by \cite{Ren2019} (Table~\ref{tab:lrat}, last row).

\begin{table}[bhtp]
\centering
\caption{\textbf{AUROC for outlier detection based on Likelihood Ratios (LRat)} (First row) LRat AUROC for outlier detection as reported in \protect\cite{Ren2019}. (Second row) LRat AUROC based on our PixelCNN++ implementation. (Third row) ``Stirring'' AUROC. (First column) FMNIST/ID vs MNIST/OOD  (Second column) CIFAR10/ID vs SVHN/OOD.}
\begin{tabular}{lrr}
\toprule
Method & FM vs MN & CF vs SV\\
\midrule
LRat (orig paper) & 99 & 93\\
LRat (our implem) & 75 & 90\\
Stirring (our method) & 98 & 94\\
\bottomrule
\end{tabular}
\label{tab:lrat}
\end{table}

\textbf{Input Complexity:} The input complexity (IC) metric  \cite{Serra2019} was computed as $S(\mathbf{x}) = -\ell_\mathcal{M}(\mathbf{x}) - L(\mathbf{x})$, where $\ell_\mathcal{M}(\mathbf{x})$ is the log likelihood (in log base 2) and $L(\mathbf{x})$ is the complexity estimates in bits for the image $\mathbf{x}$. We report results using PNG compression for the complexity estimate; these are reported as IC-PNG (Figs.~\ref{fig:grayscale} and~\ref{fig:color}, red symbols). We also estimated the score using the best complexity estimate as min$(L_1,L_2,L_3)$ using the JPEG, PNG, and FLIF compression algorithms; these are reported as IC-Best (Figs.~\ref{fig:grayscale} and~\ref{fig:color}, purple symbols).

\section{Results without conditional correction}
\label{app:without_cond}

All the results with ``stirring'' and ``shaking'' in Figures~\ref{fig:grayscale} and~\ref{fig:color} include the conditional correction as detailed in Appendix ~\ref{app:cond_corr}. Here, we show AUROC for outlier detection without conditional correction for grayscale images in Figure~\ref{app:fig:grayscale_without_cond} and for natural images in Figure~\ref{app:fig:color_without_cond}.

``Stirring'' (Fig.~\ref{app:fig:grayscale_without_cond}, orange symbols) and ``shaking'' (Fig.~\ref{app:fig:grayscale_without_cond}, green symbols) offer improvements over log likelihood (Fig.~\ref{app:fig:grayscale_without_cond}, blue symbols) in several cases while also failing in some. In particular, ``stirring'' consistently fails with Noise/OOD achieving AUROCs as low as 55\%. As pointed out in Appendix.~\ref{app:cond_corr}, Noise/OOD is a particularly challenging for a likelihood ratio as the likelihoods assigned to Noise/OOD images are very small numbers.

For natural images, ``stirring'' (Fig.~\ref{app:fig:color_without_cond}, orange symbols) and ``shaking'' (Fig.~\ref{app:fig:color_without_cond}, green symbols) perform exceptionally better than log likelihood  (Fig.~\ref{app:fig:color_without_cond}, blue symbols) even with Noise/OOD, in which case, the AUROC drops to a minimum of 88\% with ``stirring''. In summary, ``stirring'' and ``shaking'' enable robust outlier detection in many instances on their own, while conditional correction ameliorates challenges with likelihood ratio to improve performance in most failure cases.

\section{Alternative Approaches to ``Shaking''}
\label{app:alt_shaking}

\subsection{Varying patch sizes}

\begin{figure*}[!ht]
    \centering
    \begin{subfigure}[b]{0.95\textwidth}
        \centering
        \includegraphics[width=0.95\linewidth]{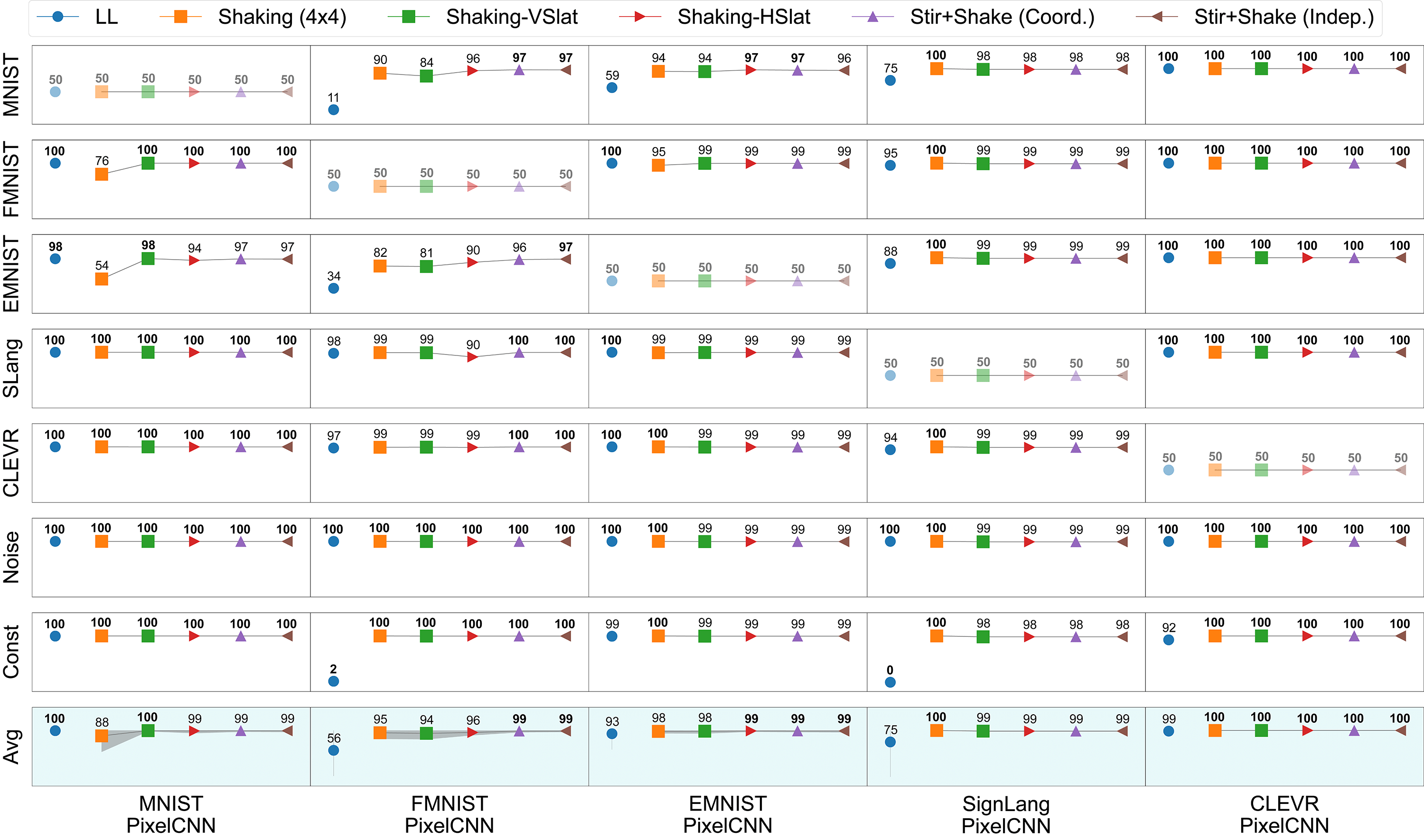}
        \caption{}
        \label{app:fig:grayscale_alt_shaking}
    \end{subfigure}
    \begin{subfigure}[b]{0.95\textwidth}
        \centering
        \includegraphics[width=0.95\linewidth]{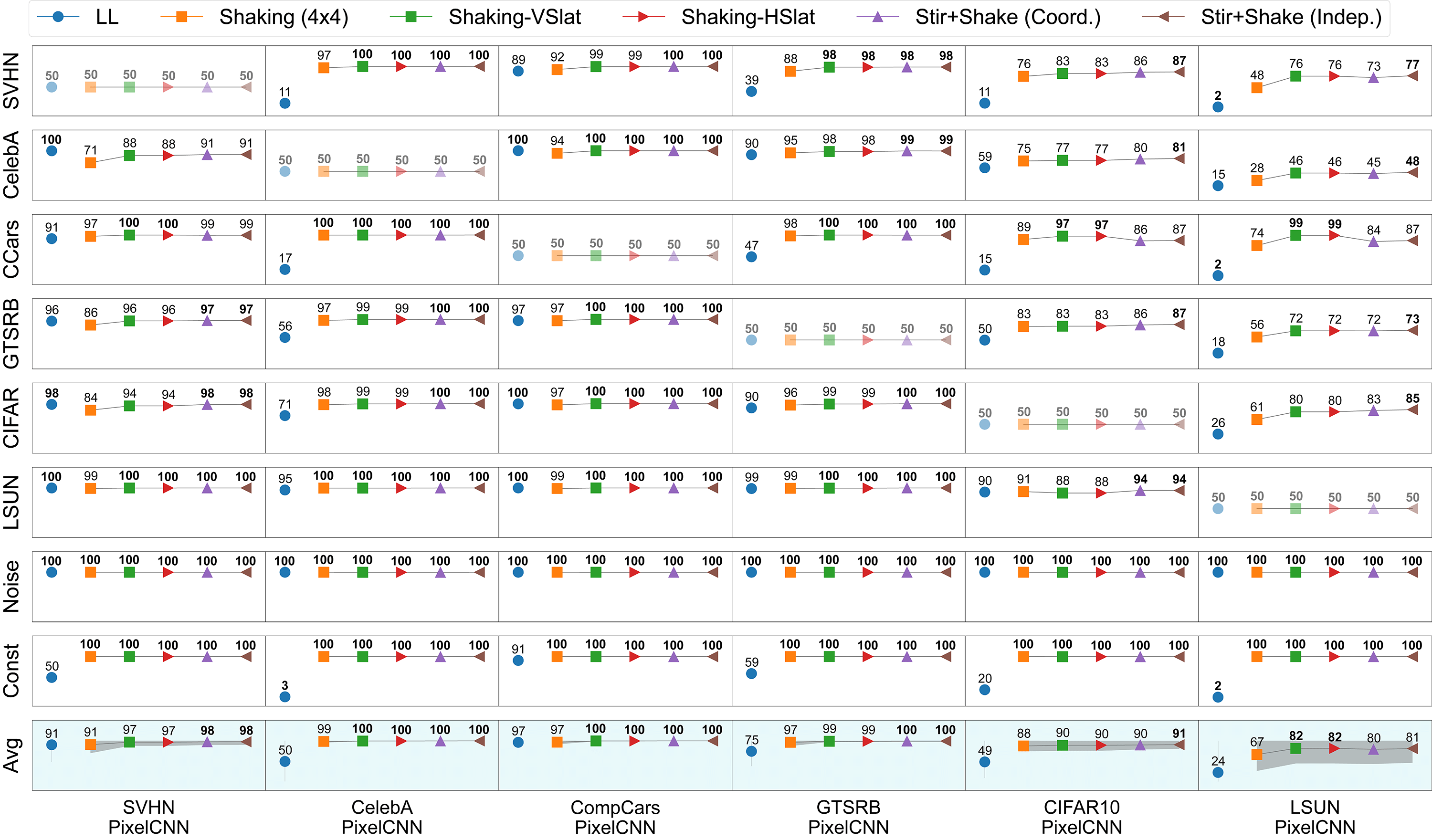}
        \caption{}
        \label{app:fig:color_alt_shaking}
    \end{subfigure}
    \caption{\textbf{Outlier detection AUROC with alternative approaches for ``shaking''.} \textbf{(a)} Grayscale image datasets. Blue: log likelihood (LL), uncorrected; Orange: Shaken LL with 16 patches; Green: Shaken LL with vertical slates; Red: Shaken LL with horizontal slats; Purple: Combination of ``stirring'' and ``shaking'' with coordinated rotation of patches; Brown: Combination of ``stirring'' and ``shaking'' with independent rotation of patches. \textbf{(b)} Same as in panel (a), but for natural image datasets. Other conventions are the same as in Figures~\ref{fig:grayscale}-\ref{fig:color}.}
    \vspace{-1em}
\end{figure*}

In the main text, we considered some simple approaches to ``shaking'', which involve splitting the image into halves or quarters and shuffling the split patches spatially. However, the are several other ways in which ``shaking'' can be performed. We describe and test a few other methods of ``shaking'' for outlier detection.

So far, we have only considered splitting the image into equal halves along the horizontal or vertical mid-lines. It is also possible to divide the image into finer segments -- either horizontally or vertically.

First, we split the images into four equally-sized vertical ``slats''. Each slat is of size 8$\times$32 pixels, which are then randomly shuffled around vertically in a coordinated manner across all 3 channels.  Second, we split the image into four equally-sized horizontal slats: four patches of size 32$\times$8 pixels. In each case, 9 derangements are possible, and we summed the change in log likelihoods across the respective derangements to compute the final outlier detection score (VSlat and HSlat, respectively).  

Lastly, we perform ``shaking'' with a finer-grained division of the image into 16 patches (4 $\times$ 4). In this case,  each patch is of size 8$\times$8 pixels. Because many derangements are possible with 16 patches, we randomly selected 20 derangements, and the final score was computed as the summed change in the log likelihoods across these 20 derangements.

We show the outlier detection performance (AUROC) based on these alternative ``shaking'' techniques in Figure~\ref{app:fig:grayscale_alt_shaking} for grayscale images and in Figure~\ref{app:fig:color_alt_shaking} for natural images (orange, green, and red symbols). We see that bigger patch sizes typically show better outlier detection scores (e.g., compare VSlat and HSlat with 4$\times$4).

\subsection{Combining ``stirring'' and ``shaking''}

We also considered combining ``stirring'' and ``shaking'' to achieve a potentially more significant disruption of long-range dependencies. We attempted two among the many different ways to combine the two approaches.

First, we consider the coordinated rotation of all patches in the image followed by shuffling. We split the image into four quarters of 16$\times$16 pixels each along the vertical and horizontal mid-lines. We then applied the same rotational transformation to all the patches in a coordinated manner, followed by spatial shuffling. Here, 63 permutations are possible (7 ``stirring'' transformations $\times$ 9 ``shaking'' derangements). The outlier detection score was computed as the summed change in the log likelihoods across 20 randomly selected derangements.

Next, we consider the independent rotation of each patch in the image, followed by shuffling. Here we also split the into four quarters of 16$\times$16 pixels but applied independent rotations to each patch, followed by spatial shuffling. As before, the outlier detection score was computed as the summed change in the log likelihoods across 20 randomly selected derangements.

The outlier detection performance (AUROC) of these combinations of ``stirring'' and ``shaking'' is shown in Figure~\ref{app:fig:grayscale_alt_shaking} for grayscale images and in Figure~\ref{app:fig:color_alt_shaking} for natural images (purple and brown symbols). We did not observe evidence for significant improvement in outlier detection by combining ``shaking'' with ``stirring'' (compare with Figs. ~\ref{fig:grayscale} and ~\ref{fig:color}, main text).

\section{Results with Medical Imaging datasets}
\label{app:medmnist_results}

\begin{figure}[h!]
    \centering
    \includegraphics[width=0.67\linewidth]{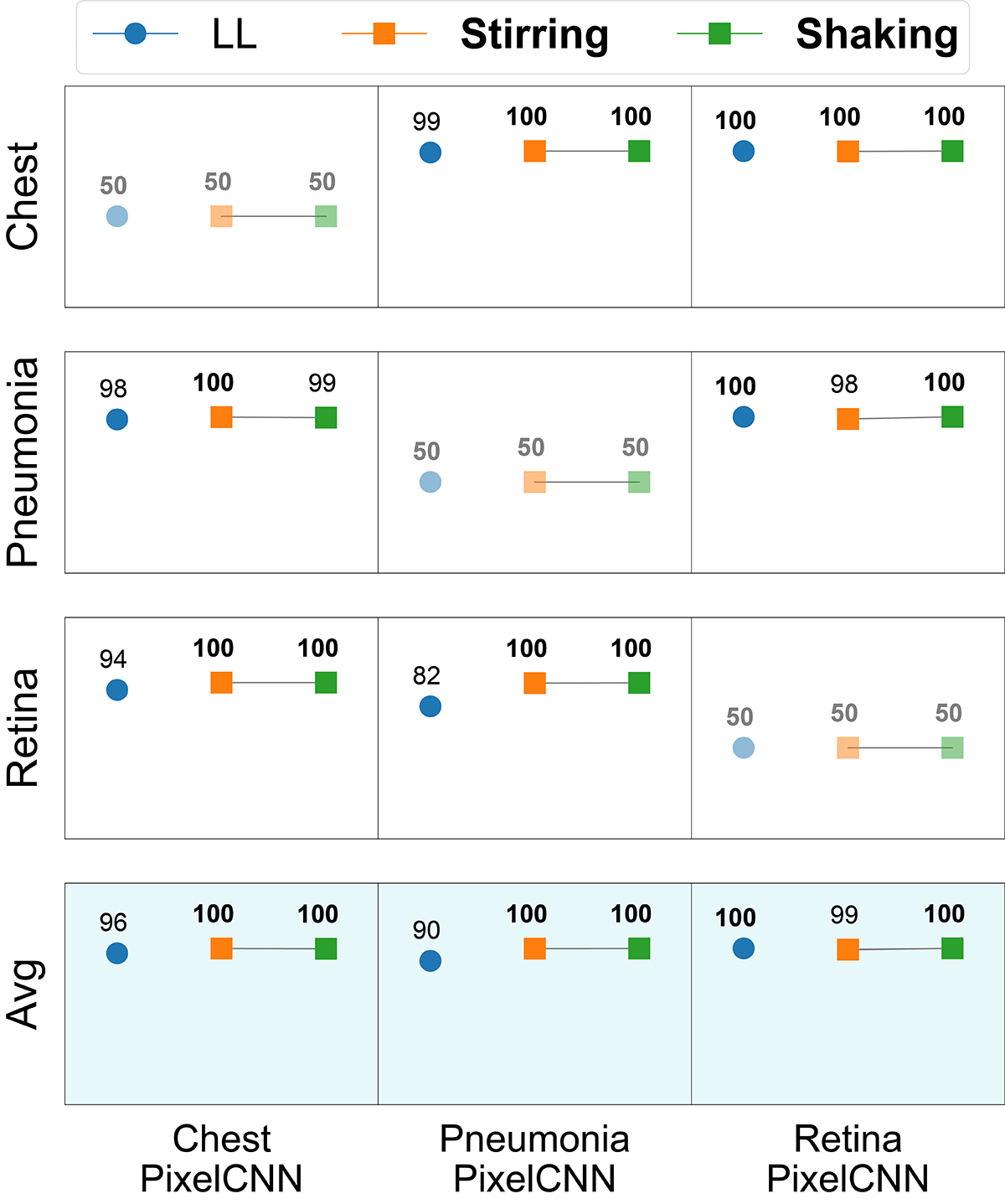}
    \caption{\textbf{Outlier detection performance in Medical Imaging.} AUROC performance for PixelCNN++ trained on MedMNIST v2 datasets. In particular, ChestMNIST, PneumoniaMNIST, and RetinaMNIST are used. Other conventions are the same as in Figure~\ref{app:fig:without_cond}.}
    \label{app:fig:medmnist}
    \vspace{-1em}
\end{figure}

In the main paper, we tested our methods extensively with standard grayscale and natural image datasets used in computer vision studies. Here, we explore other real-world applications of our methods by examining OOD detection with medical imaging datasets. 

We used a subset of medical imaging MedMNIST v2 datasets \cite{medmnistv2}, including: i) ChestMNIST -- chest X-ray images of healthy individuals and those with common thoracic diseases, ii) PneumoniaMNIST -- chest X-ray images of healthy individuals and those with Pneumonia, and iii) RetinaMNIST -- a dataset of fundus camera images of retinas for detecting diabetic retinopathy. The ChestMNIST and PneumoniaMNIST datasets can be construed as a near-OOD pair, while the RetinaMNIST was far-OOD relative to the other two datasets. In addition, RetinaMNIST images are circular fundus photographs containing several axes of symmetries, and could be a challenging case for our methods, especially ``stirring''.

We show the OOD detection performance in terms of AUROC for all pairs tested in Figure.~\ref{app:fig:medmnist}. ``Stirring'' (Fig.~\ref{app:fig:medmnist}, orange symbols) showed an average improvement of around 4\% (upto 22\%) over vanilla log-likelihood (Fig.~\ref{app:fig:medmnist}, blue symbols). Similarly, ``shaking'' (Fig.~\ref{app:fig:medmnist}, green symbols) showed an average improvement of around 4\% (upto 22\%) over log-likelihood as well, especially for the Chest and Pneumonia datasets. For RetinaMNIST, ``stirring'' and ``shaking'' did not result in any degradation in outlier detection performance despite the several axes of symmetries.

\section{Results with other generative models}
\label{app:other_gen_models}

\begin{figure*}[hbt]
    \centering
    \begin{subfigure}[b]{0.49\textwidth}
        \centering
        \includegraphics[width=0.95\linewidth]{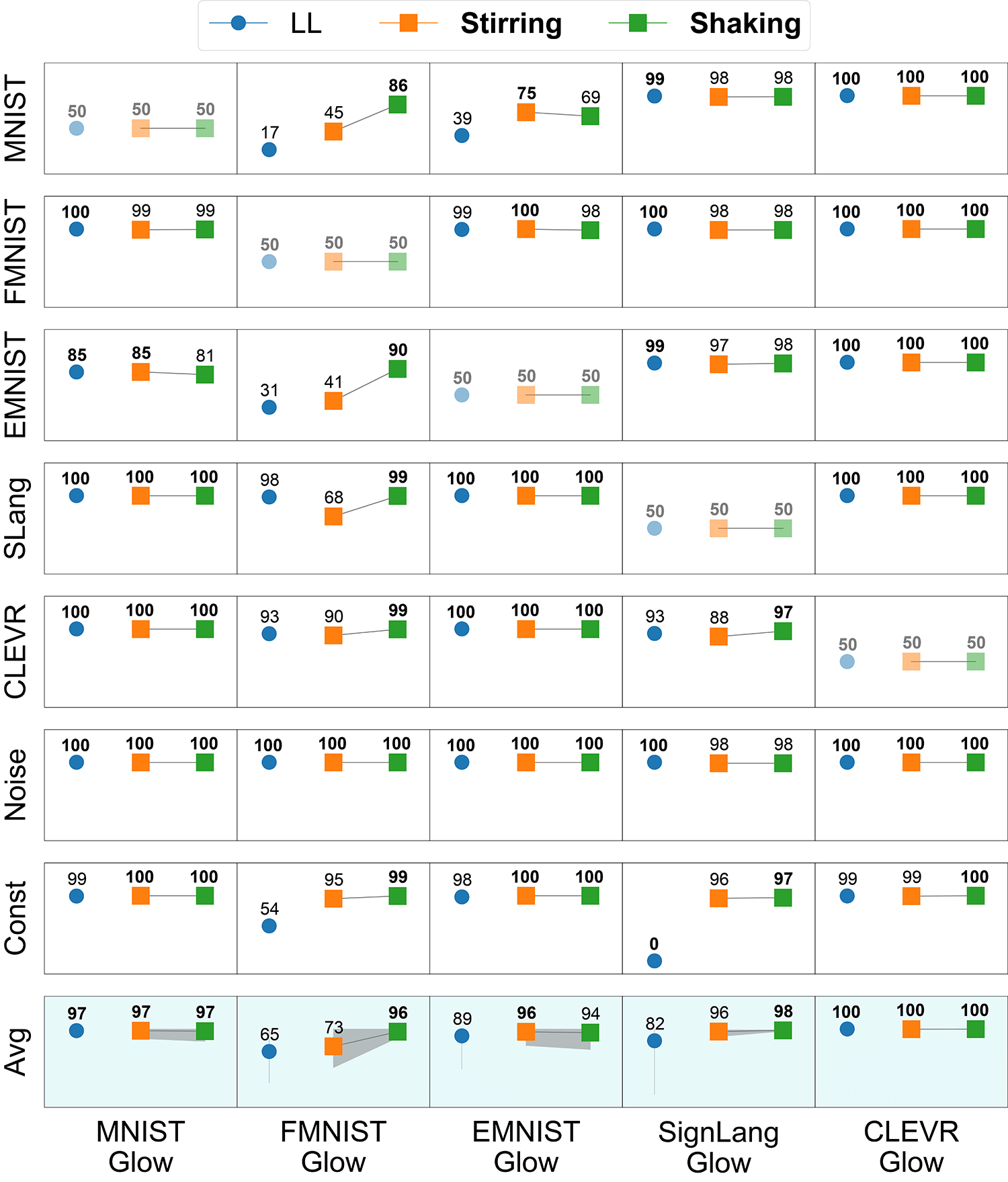}
        \caption{}
        \label{app:fig:glow_grayscale}
    \end{subfigure}
    \begin{subfigure}[b]{0.49\textwidth}
        \centering
        \includegraphics[width=0.95\linewidth]{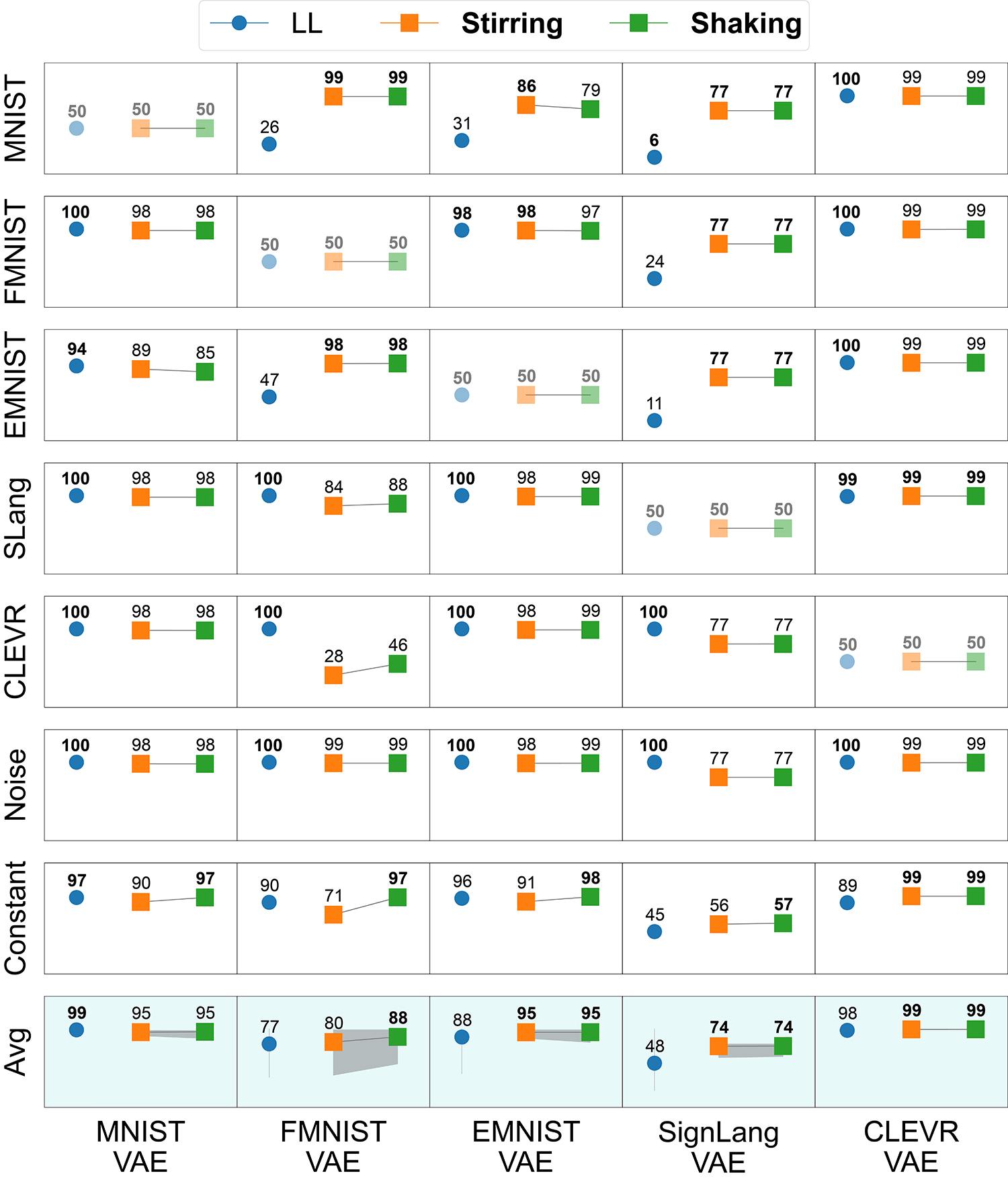}
        \caption{}
        \label{app:fig:vae_grayscale}
    \end{subfigure}
    \caption{\textbf{Outlier detection performance (AUROC) with Glow and VAE on grayscale image datasets.} \textbf{(a)} Outlier detection AUROC values for Glow models trained on grayscale image datasets. Conventions are the same as in Figure.~\ref{fig:color}.  \textbf{(b)} Outlier detection AUROC values for VAEs trained on grayscale image datasets. Conventions are the same as in Figure.~\ref{fig:color}.}
    \label{app:fig:other_gen_models_grayscale}
    \vspace{-1em}
\end{figure*}

In section.~\ref{sec:other_gen_models}, we showcased OOD detection efficacy of ``stirring'' and ``shaking'' on natural image datasets using Glow and VAE. Here we show the performance on grayscale image datasets in Figure.~\ref{app:fig:other_gen_models_grayscale}. 

For Glow models, ``stirring'' (Fig.~\ref{app:fig:glow_grayscale}, orange symbols) performed $\mathtt{\sim}$6\% better, on average, for grayscale images than vanilla log-likelihoods (Fig.~\ref{app:fig:glow_grayscale}, blue symbols). Similarly, ``shaking'' (Fig.~\ref{app:fig:glow_grayscale}, green symbols) performed $\mathtt{\sim}$11\% better, on average, for grayscale images than vanilla log-likelihoods.

For VAEs, ``stirring'' (Fig.~\ref{app:fig:glow_grayscale}, orange symbols) performed $\mathtt{\sim}$8\% better, on average, for grayscale images than vanilla log-likelihoods (Fig.~\ref{app:fig:glow_grayscale}, blue symbols). Similarly, ``shaking'' (Fig.~\ref{app:fig:glow_grayscale}, green symbols) performed $\mathtt{\sim}$10\% better, on average, for grayscale images than vanilla log-likelihoods.

\section{``Stirring'': Subpar performance with EMNIST/MNIST}
\label{app:eminst_failure}

Across the 72 ID/OOD comparisons, including both grayscale and natural image datasets, ``stirring'' outlier detection performance deviated significantly from ceiling in only one case: EMNIST ID versus MNIST OOD. We analyze this case in more detail here.

\begin{figure}[h!]
    \centering
    \includegraphics[width=0.8\linewidth]{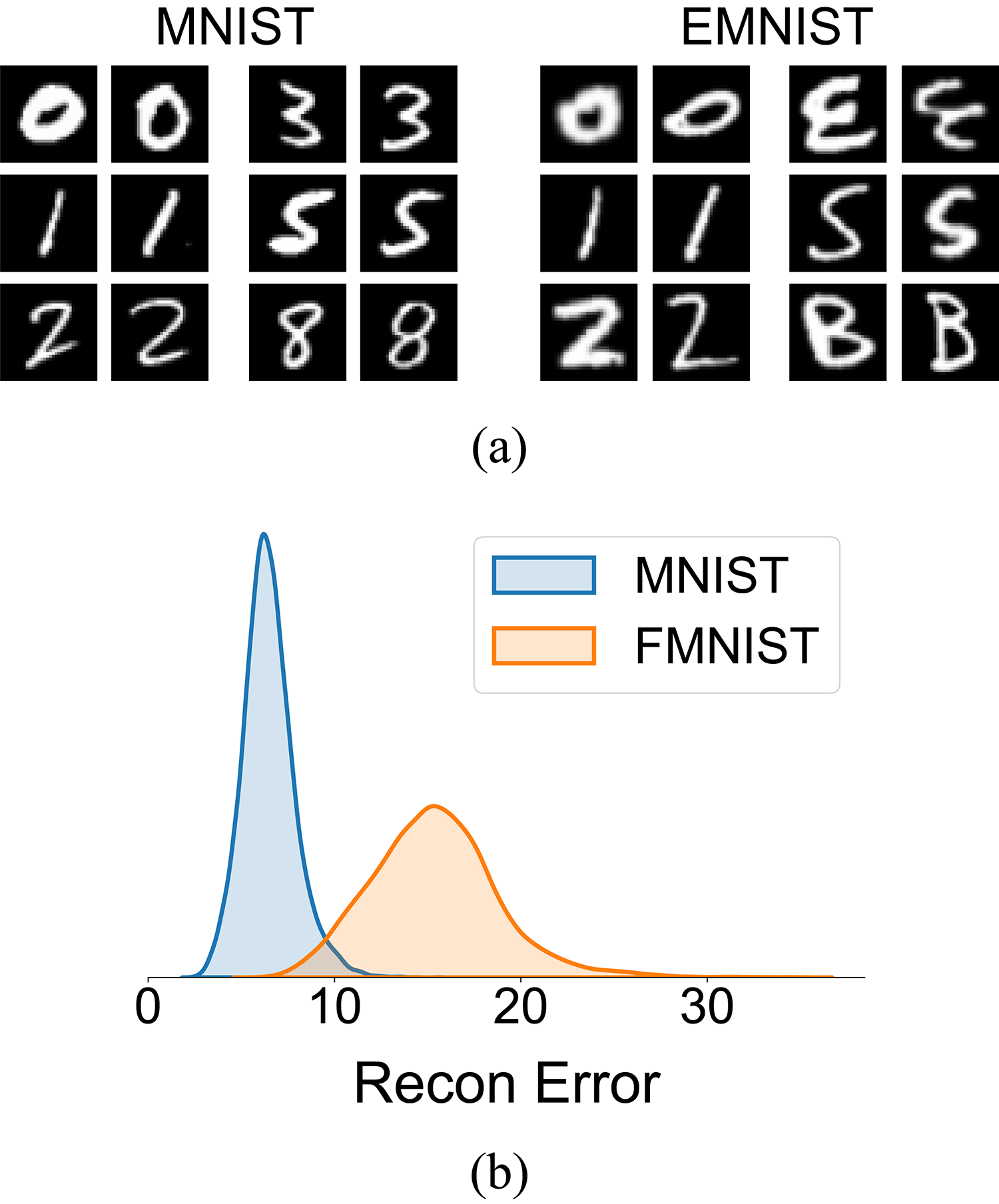}
    \caption{\textbf{(a)} Exemplars from the MNIST and the EMNIST datasets illustrate the similarity of features in the images across the two datasets. \textbf{(b)} Reconstruction error for MNIST/OOD samples (blue) and FMNIST/OOD samples (orange) for a 
    PixelCNN++ trained on the EMNIST dataset.}
    \label{app:fig:emnist_failure}
    \vspace{-1em}
\end{figure}

MNIST is a collection of hand-written numbers, and the subset of EMNIST that we used (EMNIST-letters) is a collection of hand-written English alphabets. In both these datasets, the foreground information occurs at high intensities (almost white), and the background consists primarily of low-intensity (almost black) pixels. As a result, both of these datasets comprise near-binary images. We show some exemplars of both datasets in Figure~\ref{app:fig:emnist_failure}a. It is readily observed that there are many visually similar counterparts across MNIST and EMNIST images. For example, images of 0 and O, 1 and I, 2 and Z, 3 and E, 5 and S, 8 and B look very similar as shown in Figure~\ref{app:fig:emnist_failure}a. We hypothesize that, due to this similarity, the PixelCNN++ model trained on EMNIST was powerful enough to reconstruct MNIST test images well; this yielded potential outlier detection performance. Confirming this, the reconstruction error of MNIST/OOD test samples (Fig.~\ref{app:fig:emnist_failure}b, blue density) is substantially lower than that of the FMNIST/OOD test samples (Fig.~\ref{app:fig:emnist_failure}b, orange density) for the EMNIST PixelCNN++.

\section{``Shaking'': Subpar performance with CIFAR10/LSUN}
\label{app:shaking_failure}
Across all ID/OOD comparisons, we found that ``shaking'' performed comparably with ``stirring'', by and large, in terms of outlier detection AUROC. Yet, ``shaking'' was noticeably worse than ``stirring'', particularly with more complex natural image datasets like CIFAR10 and LSUN. We speculate on the reasons here.

For two reasons, unlike SVHN, CelebA, CompCars, or GTSRB, CIFAR10 and LSUN are arguably more complicated datasets. First, these datasets are more diverse: for example, CIFAR10 comprises images of various unrelated categories, both animate (e.g., cats, deer, frogs) and inanimate (e.g., airplanes, cars, ships). Similarly, the LSUN (classroom) dataset comprises photographs of classrooms with diverse objects, including chairs, desks, chalkboards, and people. Second, images in these datasets (even within a single class) are more heterogeneous, in that key foreground objects occur at inconsistent spatial positions relative to the frame of the image. For both reasons, when split into patches across the mid-line, the likelihood that the split occurs inconsistently across key foreground objects is much higher in the more heterogeneous and diverse datasets. Consequently, it may be challenging across different images to disrupt long-range dependencies uniformly while preserving local dependencies in these datasets. We speculate that this is the primary reason for the weaker performance of ``shaking'' (as compared to shuffling) with these datasets. 

This reasoning suggests that ``shaking'' with smaller patch sizes should yield progressively worse outlier detection performance with these (CIFAR10 and LSUN) datasets, given that smaller patches would increase the chances of more inconsistent splits of foreground objects. Providing tentative validation to this hypothesis, smaller patches sizes yielded substantially worse AUROCs (upto $\sim$15\%), especially for the LSUN dataset (Fig.~\ref{app:fig:color_alt_shaking}, small 4$\times$4 patches, orange symbols versus larger slats, green and red symbols).

\section{Reproducibility Checklist}
\label{app:rep_check}

\textbf{Data:} All datasets used in this study are publicly available, and data sources are provided in Appendix~\ref{app:data}.

\noindent \textbf{Model Implementations:}
\begin{itemize}
    \item \textbf{PixelCNN++ Implementaion.} The implementational details and the selection of hyperparameters of ``tfp.distributions.PixelCNN'' to replicate our model architecture are shown in Table.~\ref{tab:arch}.
    \item \textbf{VAE Implementation.} We follow the same VAE architecture used in \cite{Chauhan_2022} (with a latent dimension of 20 and continuous Bernoulli visible distribution) and use their code which is publicly available \footnote{\url{https://github.com/google-research/google-research/tree/master/vae_ood}}.
    \item \textbf{Glow Implementation.} In our experiments, we use Glow, a type of generative flow introduced in \cite{glow}. We use TensorFlow Probability's implementation of Glow -- the ``tfp.bijectors.Glow'' class. We set the ``coupling\_bijector\_fn'' parameter to ``tfp.bijectors.GlowDefaultNetwork'', the ``exit\_bijector\_fn'' parameter to ``tfp.bijectors.GlowDefaultExitNetwork'', and ``num\_steps\_per\_block'' to 16. All other parameters are set to TensorFlow defaults. We finally construct the model using the ``tfp.distributions.TransformedDistribution'' class with the ``distribution'' set to ``tfp.distributions.MultivariateNormalDiag'' with loc 0 and scale 1, and ``bijector'' set to the Glow class.  We trained the models for 100 epochs with a batch size of 64 and a learning rate of 0.00001 using the Adam optimizer~\cite{ADAM2015}.
\end{itemize}

\noindent \textbf{Experimental results:}
\begin{itemize}
    \item Evaluation metrics (AUROC, AUPRC, FPR@80\%TPR) are standard outlier detection metrics in literature and described in Appendix~\ref{app:other_metrics}.
    \item Train/test/validation splits are detailed in Appendix~\ref{app:data}.
    \item The default shuffling orders of the test sets from their respective sources were used. For generating Noise and Constant OOD samples, NumPy\footnote{\url{https://numpy.org/}} 1.19.5 was used with a random seed of 42. The first 5000 test images were used for computing outlier detection performance.
    \item TensorFlow 2.6.0 was used to train the models, and its random seed was set to 42 to initialize the networks.
    \item We use the ``timeit'' python package for timing the code and the ``memory-profiler'' python package for profiling the code.
\end{itemize}

\noindent \textbf{Code:}
\begin{itemize}
    \item We have described, in full detail, the implementation of the $\log p_{\mathrm{LR}}$ with ``stirring'' and ``shaking'' introduced in this study (Section~\ref{sec:bijective_transformations}) for reproducibility.
    \item We have also specified in detail all the relevant software libraries and frameworks used in appropriate places (Appendices~\ref{app:architecture}, \ref{app:data}, \ref{app:other_metrics}). 
    \item Code for reproducing the results is uploaded with the Supplementary Material and will be made publicly available after paper acceptance.
\end{itemize}

\noindent \textbf{Compute Environment:} All experiments were performed on local systems equipped with two NVIDIA GTX 1080Ti GPUs and 64GB of memory.

\end{document}